\documentclass{article}

\usepackage{PRIMEarxiv}

\usepackage[utf8]{inputenc} 
\usepackage[T1]{fontenc}    
\usepackage{hyperref}       
\usepackage{url}            
\usepackage{booktabs}       
\usepackage{amsfonts}       
\usepackage{nicefrac}       
\usepackage{microtype}      
\usepackage{lipsum}
\usepackage{graphicx}
\graphicspath{{media/}}     
\usepackage{wrapfig}
\usepackage{enumitem}
\usepackage{etoc}
\usepackage{booktabs}
\usepackage{graphicx}
\usepackage{subcaption}
\usepackage{amsmath}
\usepackage{multirow}
\usepackage{cleveref}

\title{2D Spatial Reasoning with\\ Adaptive Neural Cellular Automata
}
\author{
  Martin Spitznagel$^{1}$, Janis Keuper$^{1,2}$ \\
  $^{1}$ Institute for Machine Learning and Analytics (IMLA), Offenburg University, Germany \\
  $^{2}$ University of Mannheim, Germany \\
  \texttt{firstname.lastname@hs-offenburg.de}\\\\
\textbf{Accepted at the 40th Conference on Neural Information Processing Systems \href{https://neurips.cc/}{(NeurIPS 2026)}}
}

\begin{document}
\maketitle

\begin{abstract}
Many modern learning approaches are still struggling with spatial reasoning tasks, i.e. they lack the ability to utilize geometric information of perceived entities and their spatial relation to each other to solve problems. We introduce a novel \textit{Adaptive Neural Cellular Automata} (aNCA) architecture which replaces the static and spatially invariant perception of standard NCAs by learnable, spatially variant and data-adaptive perceptive fields. We show that this conceptual change enables NCAs to iteratively reason over 2D spatial relations on grid-like data structures (e.g. images). Empirical results on public benchmarks show state of the art comprehensible results with high generalization abilities for solving image based puzzles like Sudoku or finding the shortest path in a maze.\\ \\
The implementation and all experimental setups are openly accessible as an extension to the NCAtorch framework at \url{https://github.com/mspitzna/NCAtorch}.
\end{abstract}
\vspace{-0.25cm}
\section{Introduction}
\vspace{-0.25cm}
Spatial reasoning is the ability to mentally visualize, manipulate, and functionally relate 2D and 3D objects, as well as to understand the relationships between parts, positions, and movements of one or more objects in physical space. It is a key cognitive skill used for navigation, problem-solving, engineering, and designing~\cite{kamath2023whatsup,huang2025spatialdise}. Despite its high practical impact, spatial reasoning skills are generally still underdeveloped in most current learning models~\cite{Chen_2024_CVPR, kamath2023whatsup}.\\
In contrast to the recently emerging work on \textit{``World Models''}~\cite{ding2025understanding}, which try to learn to model complex relations in a top-down fashion, we are following a bottom-up approach which tries to isolate spatial reasoning problems in controlled environments. The focus of our proposed method lies on the also open sub-problem of spatial reasoning on 2D grid structures with a fixed $x,y$-coordinate system, for which we introduce a novel architecture based on \textit{Neural Cellular Automata} ~\cite{mordvintsev2020growing}.  
Following the DISE~\cite{huang2025spatialdise} schematic of spatial reasoning taxonomies, our proposed aNCA architecture is specifically targeting 2D \textit{Extrinsic-Static} problems, that assess the relationships between multiple objects in a fixed scene (like finding the shortest path in 2D mazes~\cite{wang2025hierarchicalreasoningmodel}) and \textit{Extrinsic-Dynamic} reasoning tasks, that involve reasoning about the changing spatial relationships between
multiple objects (like solving Sudokus~\cite{wewer25srm}).

The key contributions of this paper can be summarized as follows: 
\begin{itemize}[nosep]
\item We are the first to apply the concept of learnable CA to spatial reasoning tasks, introducing \textit{Adaptive Neural Cellular Automata} (aNCA). For this purpose we move from the space- and time-invariant update rules in traditional NCA to a more powerful and flexible input-dependent, space- and time-variant regime, and show that this significantly boosts spatial reasoning performance
\item The inherent recursive nature of aNCA enables us to learn very small, yet competitive and highly generalizable models that are able to infer diverse solutions to ambiguous problems. Visualizing the learned perceptive fields allows us to make the reasoning of spatial relations interpretable. 
\item aNCA results surpass current state-of-the-art 2D spatial reasoning methods on various public benchmarks, including Sudoku, image-based Sudoku and maze shortest-path finding. 
\end{itemize}

\subsection{Related Work}
The advancement of spatial reasoning in neural networks has evolved from basic object detection to complex relational logic and artificial dynamic mental simulation. Despite the linguistic fluency of modern multimodal large language models (MLLMs), current research identifies a persistent \textit{``semantic-to-geometric gap''} where models fail to ground spatial prepositions in coherent geometric world models \cite{Chen_2024_CVPR, kamath2023whatsup}.\\
\noindent\textbf{Cognitive Frameworks and Taxonomies.}
To diagnose these systematic failures, recent literature has adopted cognitively grounded taxonomies. The DISE framework \cite{huang2025spatialdise} categorizes spatial tasks into a $2 \times 2$ matrix based on Intrinsic vs. Extrinsic properties and Static vs. Dynamic environments. While models are proficient in static settings, they exhibit a high failure rate in tasks requiring dynamic mental simulation, primarily due to deficits in ``spatial working memory'' and an inability to apply basic geometric axioms.\\
\noindent\textbf{Architectural Inductive Biases.}
Traditional architectures often lack the structure for relational reasoning. To address this, Relation Networks (RNs) were introduced to explicitly compute pairwise relations between all objects in a set \cite{santoro2017simple}. For more complex scenes, Scene Graph Neural Networks (SGNNs) represent environments as structured graphs where nodes denote objects and edges represent their relationships, utilizing message-passing to propagate context \cite{zhu2022knowledge, pham2025tesgnn}. More recently, mechanisms for Vision Transformers (ViTs), such as Context-Aware Gating (CAG) in the Spatial Decay Transformer (SDT), inject data-dependent spatial decay into self-attention to selectively focus on semantically relevant regions while maintaining a locality bias \cite{mao2026learning, fan2024rmt}.\\
\noindent\textbf{General Spatial Benchmarking.}
No generally accepted closed-form definition reduces spatial reasoning ability to a single numerical score. Evaluation therefore relies on proxy problems that isolate particular spatial capabilities; designing such benchmarks is itself an active field of research, and their results provide task-specific rather than universal measures~\cite{huang2025spatialdise}. Benchmarking spatial reasoning abilities has transitioned from early synthetic 2D/3D datasets like CLEVR \cite{Johnson_2017_CVPR} and GQA \cite{hudson2019gqa} to complex 3D and multi-hop scenarios. Benchmarks such as StepGame \cite{shi2022stepgame} and SpartQA \cite{mirzaee-etal-2021-spartqa} test multi-step logical inference in text and vision. For 3D awareness, 3DSRBench \cite{ma20253dsrbench} and Ego3D-Bench evaluate robustness across varying camera viewpoints, while SITE \cite{wang2025site} and OmniSpatial \cite{jia2025omnispatial} provide comprehensive evaluations across diverse spatial factors including figural and environmental scales. SpatiaLQA emphasizes ``spatial logical reasoning'' within open-vocabulary indoor environments \cite{xie2026spatialqa}.\\
\noindent\textbf{Spatial Reasoning on 2D (Grid) Data.} While most of the previous benchmarks focus on spatial reasoning in 3D scenes (or 3D$\rightarrow$2D projections of these), there is a growing body of recent works that focus on pure 2D spatial reasoning: e.g. ~\cite{wewer25srm} introduced spatial reasoning with SRN Diffusion Models which operate on images. Another line of work proposed to solve 2D reasoning with \textit{Hierarchical Reasoning Models }(HRM)~\cite{wang2025hierarchicalreasoningmodel}, an approach which has recently been extended by ItrSA++~\cite{kubo2026cvoting} and TRM~\cite{jolicoeurmartineau2025morerecursivereasoningtiny}.\\
\noindent\textbf{Benchmarks for 2D Spatial Reasoning.} 2D tasks not only provide simpler spatial reasoning problems which allows the study of novel approaches more easily, but also have several practically relevant benchmark applications like solving Sudoku on images of handwritten digits~\cite{wewer25srm}, finding the shortest path in a maze~\cite{wang2025hierarchicalreasoningmodel} or finding spatially local solutions with global constraints~\cite{wewer25srm}.\\
We compare aNCA with state-of-the-art SRM~\cite{wewer25srm}, HRM~\cite{wang2025hierarchicalreasoningmodel}, and TRM~\cite{jolicoeurmartineau2025morerecursivereasoningtiny} on the tasks introduced in \cref{sec:spatial_reasoning}.

\section{Neural Cellular Automata} \label{sec:methods}
\begin{wrapfigure}{r}{0.5\textwidth}
\vspace{-0.5cm}
    \centering
        \includegraphics[scale=0.2]{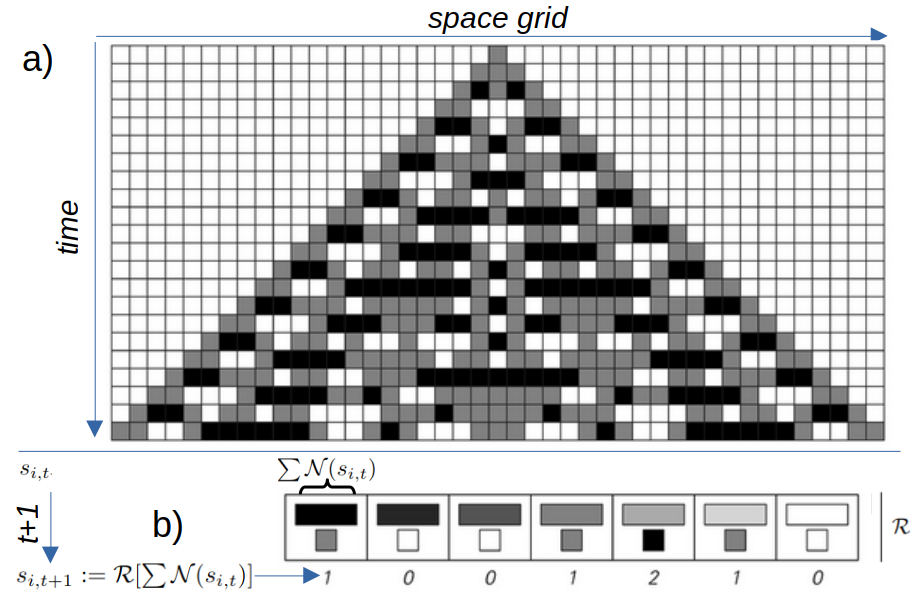} 
    \caption{Sketch of a basic 1D Cellular Automaton with three discrete states $s_i \in \{0,1,2\}$ and seven update rules $\mathcal{R}$, which operate on the sum over a 3-cell neighborhood $\mathcal{N}$. a) shows the inference of complex patterns from an initial state by the rules visualized in b) over time. }
\label{fig:basicCA}
\end{wrapfigure}
Originating from the seminal work of \textit{John von Neumann}~\cite{von1966theory}, \textit{Cellular Automata} (CA) are discrete computational systems studied since the 1940s. Typically, a CA is defined on a regular, finite grid $\mathcal{G}$ composed of discrete cells. At each time step $t$, every cell $i$ maintains a discrete state $s_{i,t}$. The system's evolution is governed by the synchronous application of a rule set $\mathcal{R}$ across all cells, where the subsequent state $s_{i,t+1}$ is derived from the local neighborhood $\mathcal{N}(s_{i,t})$ such that $s_{i,t+1}:=\mathcal{R}[\mathcal{N}(s_{i,t})]$~\cite{schiff2011cellular}. This process is illustrated in \Cref{fig:basicCA}.

CAs gained significant prominence in the 1970s via \textit{Conway's ``Game of Life''}~\cite{conway1970game}, a 2D CA utilizing binary states $s_{(x,y),t} \in \{0,1\}$ and $3\times 3$ neighborhood rules. CAs have proven effective in modeling diverse phenomena in biology~\cite{coombes2009geometry, bouligand1986fibroblasts, hatzikirou2012go}, chemistry~\cite{gerhardt1989cellular}, and physics~\cite{wolfram1983statistical, zaluska2021step}. Furthermore, their status as a universal model of computation was confirmed by the proof of \textit{Turing Completeness}~\cite{cook2004universality} for specific configurations. These developments underpinned \textit{Stephen Wolfram's} \textit{``A New Kind Of Science''}~\cite{wolfram2003new}, which posited that complex scientific problems might be better characterized by simple recursive programs like CAs than with differential equations.\\
Traditional CA models, such as those discussed in~\cite{wolfram2003new},~\cite{conway1970game}, and~\cite{von1966theory}, utilize manually defined rule sets. While these rules can be enumerated combinatorially for a given state space $\mathcal{S}$ and neighborhood size $\mathcal{|N|}$, the search space expands exponentially; for instance, a binary 2D CA with a $3\times 3$ neighborhood already permits $2^{2^{3 \times 3}}=2^{512}$ possible rules. To address this complexity, \cite{mordvintsev2020growing} proposed learning rules from data, introducing \textbf{\textit{Neural Cellular Automata} (NCA)}. In this framework, handcrafted rules are replaced by an artificial neural network $f_\phi$ with learnable parameters $\phi$. The update step is expressed as:
\begin{equation}
s_{i,t+1}:=s_{i,t}+f_\phi[\mathcal{N}(s_{i,t})]
\label{eq:nn_update}
\end{equation}

\begin{wrapfigure}{r}{0.5\textwidth}
    \centering
        \includegraphics[scale=0.4]{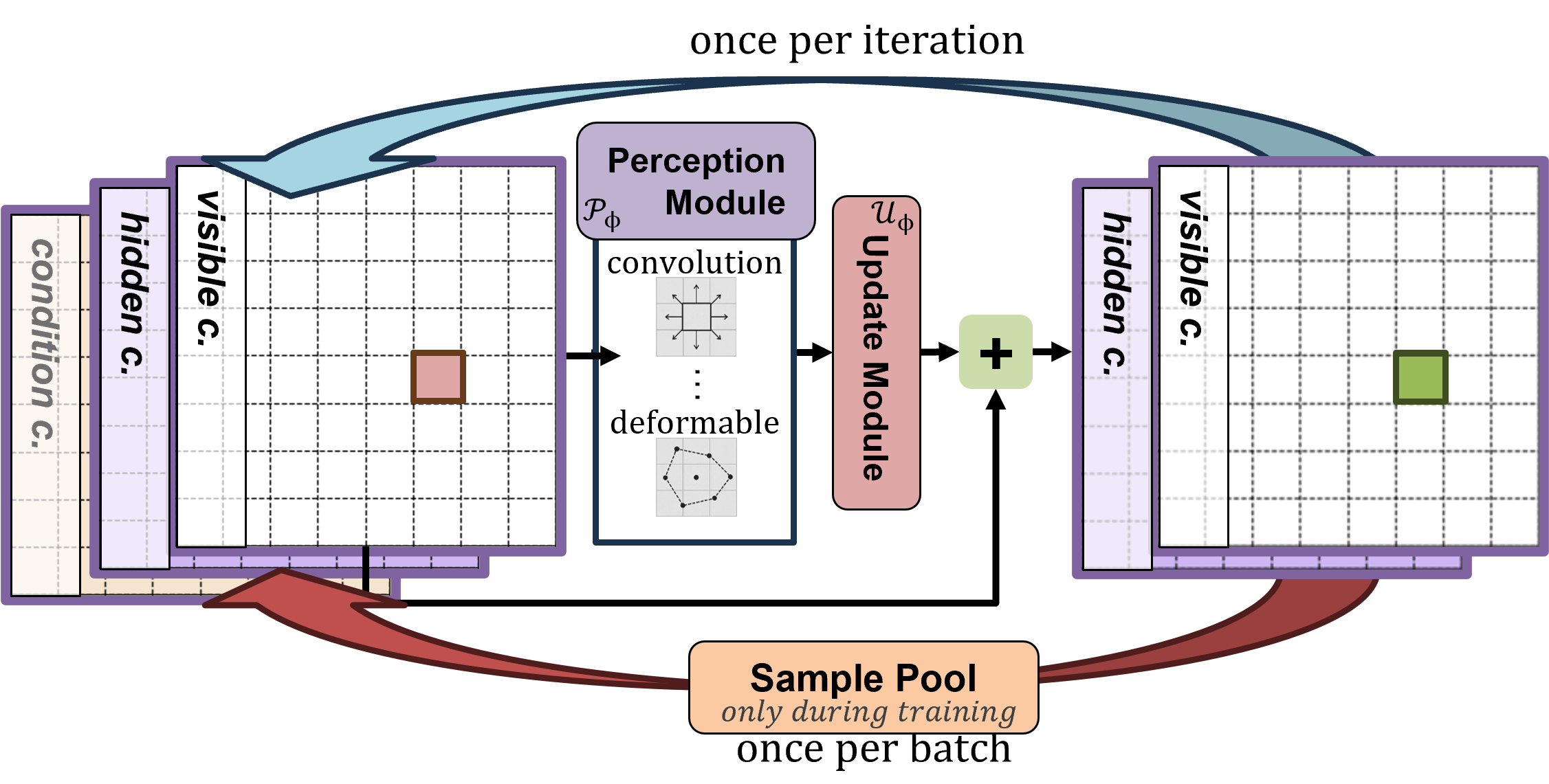} 
    \caption{Sketch of a basic CNN implementation~\cite{spitznagel2026newkindnetworkreview} of a 2D NCA with a 3D state space. The perception module processes the cell state using one of several configurable perception options. The update module processes the perceived features to compute state updates. Stochastic updates are applied before updating the visible and hidden channels to form the new cell state for the next iteration. The sample pool is used after one batch of images is processed to retain and reuse evolved states for subsequent training iterations.}
\label{fig:basicNCA}
\end{wrapfigure}

Since $f_\phi$ is applied identically to every position $i$ (spatial invariance), \cite{mordvintsev2020growing} utilized ``nested'' \textit{Convolutional Neural Networks} (CNNs)~\cite{lecun1998convolutional} for efficient implementation. Here, convolutional filters function as learnable local rules over kernel-defined neighborhoods. This architecture is depicted in \Cref{fig:basicNCA}. While \cite{gilpin2019cellular} established the theoretical equivalence between CA and such CNNs, most NCA models depart from tradition by employing continuous vector-valued states $s_{i,t}\in\mathbb{R}^n$ rather than discrete ones.

Initial work by \cite{mordvintsev2020growing} demonstrated robust self-organization and classification using single-layer convolutional NCAs. This has since expanded into more expressive architectures: \cite{sandler2020image} applied larger CNNs for segmentation, \cite{tesfaldet2022attention} introduced attention-based variants and \cite{10204488} real-time dynamic texture synthesis using NCAs. Furthermore, NCAs have been integrated with GANs \cite{otte2021generative}, VARs \cite{palm2022variational}, and diffusion models \cite{kalkhof2024frequency} for generative tasks, while \cite{grattarola2021learning} extended the paradigm to graph-structured data. Due to their inherent spatial structure, NCAs intuitively appear to be well suited for (2D) spatial reasoning tasks. However, to the best of our knowledge, this has never been attempted before.

\subsection{NCA Perception Modules}
Building on \cref{eq:nn_update}, we follow the notation from~\cite{spitznagel2026newkindnetworkreview} and factor the learnable update rule $f_\phi$ into two distinct sub-modules: a \textit{perception module} $\mathcal{P}_\phi$ and an \textit{update module} $\mathcal{U}_\phi$, and re-formulate the NCA iteration as:
\begin{equation}
s_{i, t+1} = s_{i, t} + \mathcal{U}_\phi\bigl[\mathcal{P}_\phi(s_{i, t})\bigr].
\label{eq:decomposed_update}
\end{equation}
The perception module $\mathcal{P}_\phi$ determines how each cell reads information from its neighborhood and therefore controls the effective receptive field and communication structure of the automaton, while the update module $\mathcal{U}_\phi$ transforms the perceived features into the state increment applied to the cell. In our implementation, the cell state is first concatenated channel-wise with the task condition and, when enabled, fixed positional embeddings. The resulting tensor is processed by one or more perception branches operating in parallel; their outputs are concatenated and passed to a shared $1\times 1$ convolutional update network that predicts the state increment. Because we keep $\mathcal{U}_\phi$, the cell-state structure, and the training protocol fixed across experiments, any difference in task performance can be attributed to the inductive bias of $\mathcal{P}_\phi$. \Cref{fig:basicNCA} illustrates how these components interact within a single NCA step. In the remainder of this section we describe three perception variants: standard $k\times k$ convolution (e.g. $3\times 3$ or $5\times 5$), dilated convolution and a Sudoku-specific constraint-aligned variant that combines row, column, and $3\times 3$-box convolutions.\\
%
%
\noindent\textbf{Standard $k\times k$ convolution.} Our baseline perception is a single learnable $k\times k$ convolution layer with 128 kernels and leaky-ReLU activation and circular padding, following the original formulation of~\cite{mordvintsev2020growing}. Each cell's receptive field is therefore limited to its immediate neighbors per step, and longer-range dependencies must be built up recurrently.\\\\
\noindent\textbf{Dilated Convolutions.}
We propose dilated convolutions~\cite{yu2016multiscalecontextaggregationdilated} as an NCA perception module for spatial reasoning tasks requiring long-range communication. A dilated $3\times 3$ kernel with rate $r$ samples a $(2r+1)\times(2r+1)$ grid region at no extra parameter cost, directly enlarging the per-step receptive field. In the NCA setting, fewer iterations are needed to propagate information across the grid, which is a particularly effective inductive bias for tasks where constraints span the entire grid.\\
\noindent\textbf{Manually Constraint-Aligned Convolutions.}
For tasks whose constraint structure is known a priori, we propose to integrate that structure manually into the perception module itself rather than asking the NCA to discover it over many iterations. For Sudoku in particular, every valid solution must satisfy three constraint families, one per row, column, and $3\times 3$ box, each of which spans an entire "unit" of the $9\times 9$ grid. We therefore combine three parallel perception branches whose kernels match the shapes of these units: a $1\times 9$ \textit{row convolution}, a $9\times 1$ \textit{column convolution}, and a structured $3\times 3$ \textit{box perception} that aggregates the other cells in each $3\times 3$ box.
\vspace{-0.25cm}
\subsection{Training NCA}
\begin{wrapfigure}{r}{0.5\textwidth}
\vspace{-0.5cm}
  \centering
  \includegraphics[width=1\linewidth]{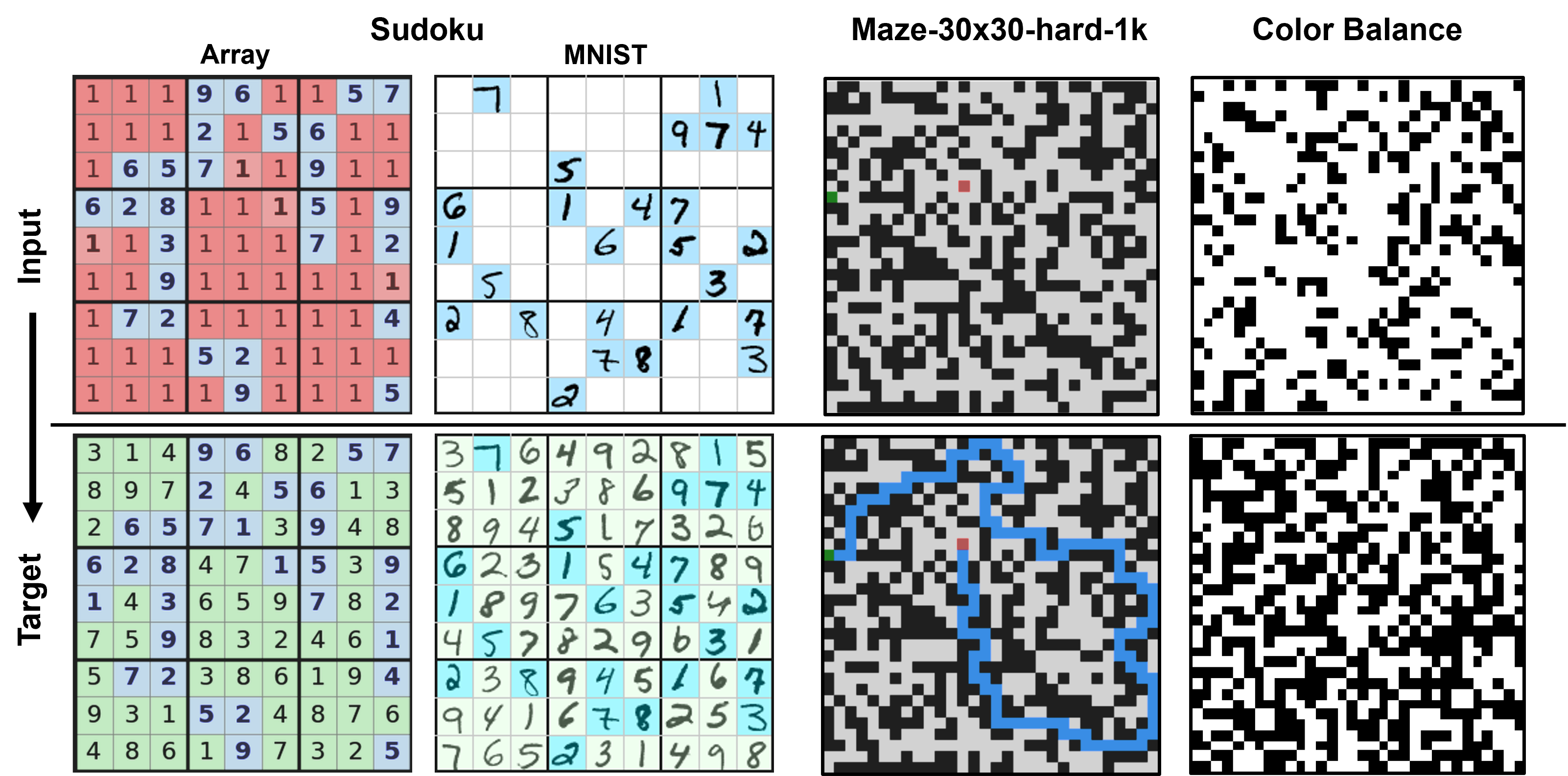}
  \caption{The four 2D spatial reasoning tasks evaluated in this work. Each column shows the input (top) and target (bottom).}
  \label{fig:tasks}
  \vspace{-0.25cm}
\end{wrapfigure}
Following~\cite{spitznagel2026newkindnetworkreview}, we train the NCA by unrolling the update rule for a randomly sampled number of steps $T\sim\mathcal{U}[T_\text{min}, T_\text{max}]$ and comparing the resulting state to the target output under a task-specific objective. For conditional tasks such as \textit{Sudoku} and maze solving, each cell is initialized from the puzzle input and conditioned on the same input throughout the rollout, so that the observable evidence remains available at every step. Gradients are backpropagated through the full unrolled trajectory. To further encourage temporal stability, we adopt the \textit{sample pool} strategy of~\cite{mordvintsev2020growing}: a fraction of each mini-batch is initialized from previously evolved states drawn from a fixed-size buffer rather than from fresh seeds, and the newly evolved states are written back into the pool after each training step. This persistent-state regime exposes the update rule to intermediate configurations far from the initial seed and discourages it from collapsing or diverging on longer rollouts, which is essential for tasks that require many iterations to propagate information across the grid. Full hyperparameter details for each task are listed in \cref{app:hyperparameters_sudoku,app:hyperparameters_maze,app:hyperparameters_color_balance} in the Appendix.
\vspace{-0.25cm}
\section{Spatial Reasoning with NCA} \label{sec:spatial_reasoning}

To the best of our knowledge, we are the first to propose the use of NCAs for (2D) spatial reasoning tasks. We cast each spatial reasoning task as a recurrent computation on a 2D grid $\mathcal{G}$, solved by the NCA defined in \cref{eq:nn_update}. All tasks share the same architecture, optimizer, and sample-pool training regime: we unroll for $T\sim\mathcal{U}[T_\text{min}, T_\text{max}]$ steps, optimize with AdamW under gradient clipping, and concatenate the condition channel-wise into the cell state at \emph{every} step. Because high-difficulty Sudoku puzzles admit many valid completions, path-finding admits shortest-path ties, and the color-balance task has no per-pixel supervision, we use \emph{constraint-based} objectives that reward any valid output. The four tasks, illustrated in \cref{fig:tasks}, use established benchmarks from prior work.

\begin{figure}[h!]
  \centering
  \includegraphics[width=0.7\linewidth]{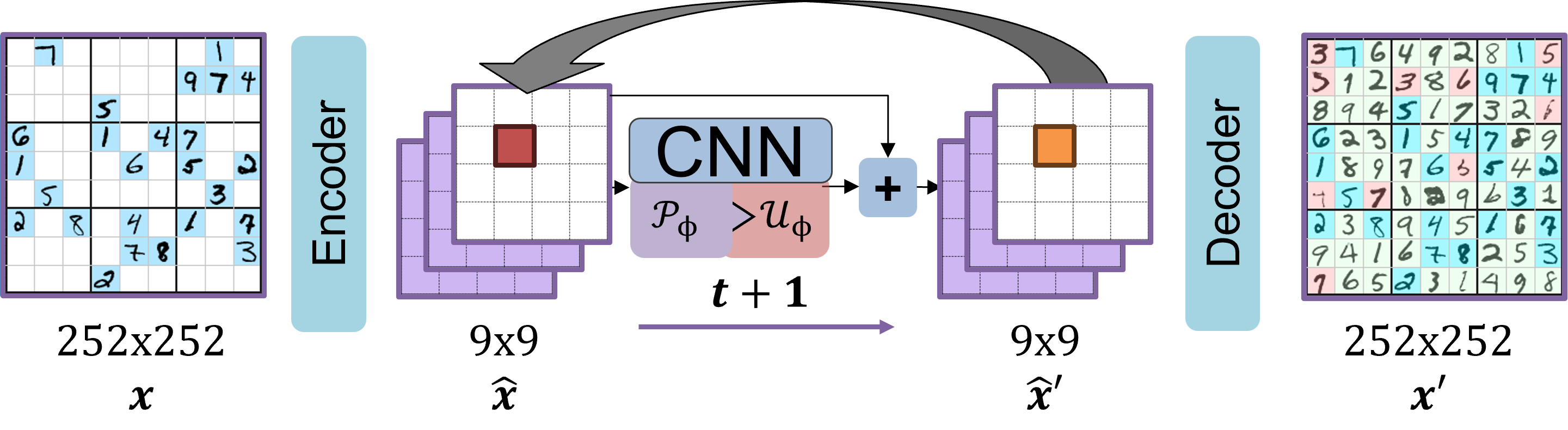}
  \caption{Latent pixel Sudoku pipeline. The encoder $E_\psi$ maps the initial handwritten Sudoku image into a $9\times 9\times 9$ grid of digit logits, the NCA iterates in this latent board space under the Sudoku constraint loss until a valid completion emerges, and the decoder $D_\omega$ renders the final board back to pixels. Freezing $E_\psi$ and $D_\omega$ decouples perception from reasoning and allows different NCA variants to be evaluated within the same image-space pipeline.}
  \label{fig:pixel_sudoku_nca}
  \vspace{-0.25cm}
\end{figure}

\noindent\textbf{Sudoku.} We use the one-million-puzzle dataset of~\cite{wewer25srm}, generating puzzles on the fly by masking cells at a configurable difficulty; a held-out set of solutions is never seen during training. Each cell carries $D=9$ visible digit channels and the condition $c_i\in\{0,1\}$ marks clue cells. The objective exploits the puzzle's constraint structure: letting $p_{i,d}$ be the softmax probability of digit $d$ at cell $i$, we penalize the squared deviation of per-digit unit sums from one across all rows, columns, and $3\times 3$ boxes, rewarding any structurally valid completion independently of which specific digits the NCA commits to. Training hyperparameters, loss details, and computational cost are provided in \cref{app:sudoku}.\\
\noindent\textbf{MNIST Sudoku.} We reuse the same solutions and split, rendering each cell as a handwritten glyph drawn from MNIST to give a $9s\times 9s$ image where each $s\times s$ tile is blank or carries a digit. The condition is a binary pixel mask over given cells. End-to-end NCA training exposed a trade-off: rollouts either produced crisp MNIST glyphs without satisfying Sudoku constraints, or satisfied constraints while letting glyphs drift. We therefore train a separate \textit{Sudoku codec} (\cref{fig:pixel_sudoku_nca}): a frozen encoder $E_\psi$ maps a pixel Sudoku to a $9\times 9\times 9$ digit-logit grid and a frozen decoder $D_\omega$ renders a board back to pixels via a conditional VAE with a style latent. The NCA operates entirely in board latent space under the same constraint objective as the array variant, with per-digit probabilities supplied by a frozen MNIST-CNN applied to the decoded tiles rather than read from raw digit channels. Codec architecture, training details, and an isolation evaluation of encoder and decoder accuracy are provided in \cref{app:mnist_sudoku}.

\noindent\textbf{Maze solving.} We use the $30\times 30$ maze dataset of~\cite{wang2025hierarchicalreasoningmodel} (1K train, 1K test), with shortest-path targets. The condition encodes walls, start, and goal. The target is a binary shortest-path mask. Since walls are trivially predictable from the condition, the loss is restricted to non-wall cells and combines a Dice term (handling path/non-path imbalance), a BCE term (encouraging binary commitment), and a normalized length penalty $\lvert\text{len}_\text{pred}-\text{len}_\text{true}\rvert$. Training hyperparameters and loss details are provided in \cref{app:maze}.\\
\noindent\textbf{Global color balance.} We adapt the global-constraint task of~\cite{wewer25srm} to black-and-white pixels. Starting from a randomly biased binary image, the NCA must fill an unconditioned grid to realize a prescribed black/white ratio (e.g.\ $50/50$), with no per-pixel supervision. The objective combines a count term pushing the global color mean toward the target ratio and a saturation term that zeroes only at fully committed pixel values. Training hyperparameters and loss details are provided in \cref{app:color_balance}.
\vspace{-0.25cm}
\section{Adaptive Neural Cellular Automata (aNCA)}
\begin{figure}[h!]
  \centering
  \begin{minipage}{0.8\linewidth}
    \centering
    \begin{subfigure}[t]{0.26\linewidth}
      \centering
      \includegraphics[width=\linewidth]{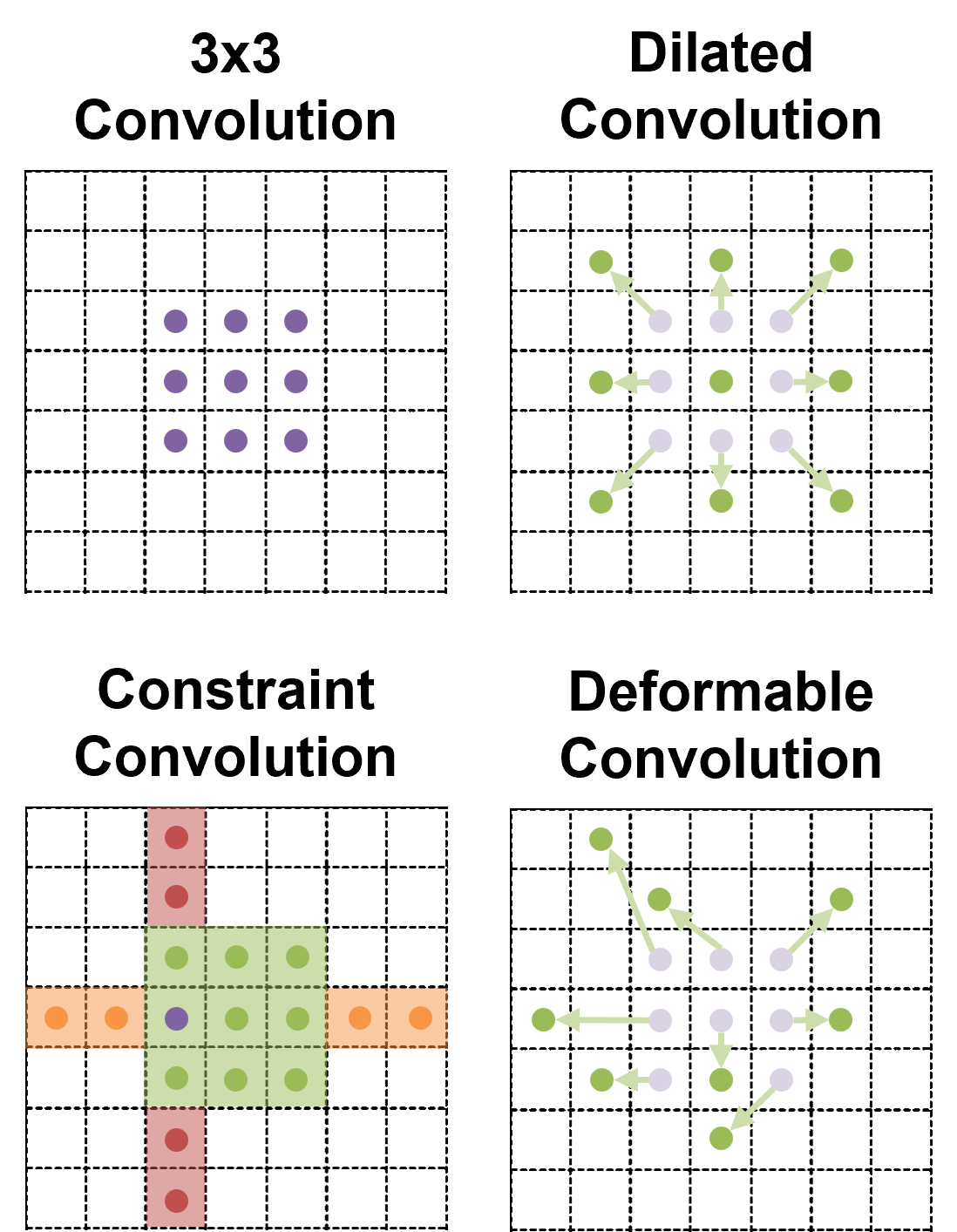}
      \caption{Fixed vs. adaptive perception patterns.}
      \label{fig:anca_perceptions}
    \end{subfigure}
    \hfill
    \begin{subfigure}[t]{0.73\linewidth}
      \centering
      \includegraphics[width=\linewidth]{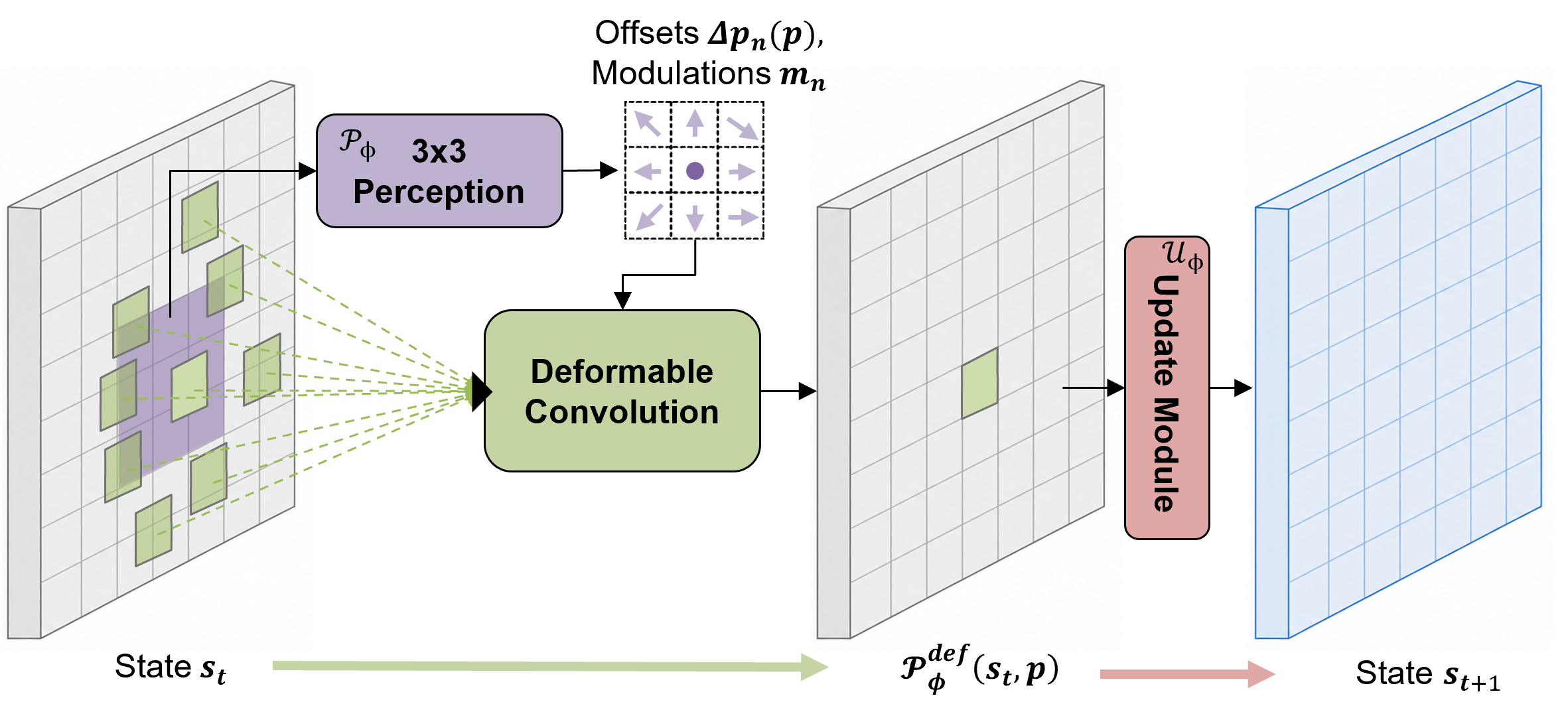}
      \caption{aNCA single-step computation graph.}
      \label{fig:anca_diagram}
    \end{subfigure}
  \end{minipage}
  \caption{(a) Comparison of perception sampling patterns: standard $3\times3$, dilated, constraint-aligned, and deformable convolution. (b) One aNCA update step: the perception module $\mathcal{P}_\phi$ predicts offsets $\Delta\mathbf{p}_n(\mathbf{p})$ and modulations $m_n$ from $s_t$, which guide the deformable convolution to aggregate features from adaptive grid locations. The update module $\mathcal{U}_\phi$ then produces the next state $s_{t+1}$.}
  \label{fig:anca}
\end{figure}

We propose \textit{Adaptive Neural Cellular Automata} (aNCA), which extend the reasoning abilities from learned space- and time-invariant update rules of traditional NCA to a more powerful and flexible input-dependent, space- and time-variant regime. We achieve this significant change in NCA capabilities through a multi-head perception module based on DCNv2 \textit{deformable convolutions}~\cite{dai2017deformableconvolutionalnetworks, DBLP:journals/corr/abs-1811-11168}, which instantiates the NCA framework of \cref{eq:decomposed_update} as perception module $\mathcal{P}_\phi$ and allows the effective neighborhood $\mathcal{N}$ to adapt across grid positions and NCA steps. \Cref{fig:anca} illustrates the architecture: \cref{fig:anca_perceptions} contrasts the fixed sampling patterns of the perception variants above against the adaptive sampling of the proposed deformable perception, \cref{fig:anca_diagram} shows the full single-step aNCA computation graph. The deformable perception module is formalized below.

\noindent\textbf{Adaptive Perception via Deformable Convolutions.}
Unlike dilated convolutions, which fix the neighborhood geometry at initialization, \textit{deformable convolutions}~\cite{dai2017deformableconvolutionalnetworks} augment each kernel tap with a learned, position-dependent spatial offset. Rather than committing to a fixed sampling pattern, the model learns \emph{which} positions each cell should attend to, depending on the current input. This allows it to discover long-range constraint structures from data without explicit apriori modeling of these structures. Concretely, a standard $k\times k$ convolution at grid position $\mathbf{p}$ reads
\begin{equation}
  \mathcal{P}(s,\mathbf{p})
    = \sum_{n=1}^{k^{2}} w_n \cdot s\!\left(\mathbf{p}+\mathbf{p}_n\right),
\end{equation}
where $\mathbf{p}_n$ enumerates the $k^2$ fixed kernel positions and $w_n$ are learnable weights. We use the DCNv2 variant~\cite{DBLP:journals/corr/abs-1811-11168}, which augments each kernel position with both a spatial offset $\Delta\mathbf{p}_n(\mathbf{p})\in\mathbb{R}^2$ and a learned modulation scalar $m_n(\mathbf{p})\in[0,2]$, each predicted by a separate auxiliary convolution applied to $s_{i,t}$. Its difference from the offset-only DCNv1 variant is detailed in \cref{app:dcn_comparison}. Fractional positions are resolved by bilinear interpolation:
\begin{equation}
  \mathcal{P}_\phi^{\mathrm{def}}(s,\mathbf{p})
    = \sum_{n=1}^{k^{2}} w_n \cdot m_n(\mathbf{p})
      \cdot s\!\left(\mathbf{p}+\mathbf{p}_n+\Delta\mathbf{p}_n(\mathbf{p})\right).
  \label{eq:deform_conv}
\end{equation}
The modulation scalar allows each kernel position to be suppressed or amplified independently, giving the model additional flexibility beyond the spatial offset alone. We stack $H=3$ deformable branches in parallel as $\mathcal{P}_\phi$, producing $3\times(2+1)\times 9=81$ offset and modulation values per cell per step. Different heads can thereby specialize to complementary constraint directions. Because the learned offsets are explicit and inspectable, we can directly audit which spatial relations the NCA has discovered. In \Cref{sec:visualizing_spatial_relations} we show that the three heads learn sampling patterns that align with Sudoku's row, column an box constraints, without encoding this explicitly in the architecture.\\
Our approach significantly differs from the "deformable convolution" implemented in in the NCATorch framework of~\cite{spitznagel2026newkindnetworkreview} - there a single deformable convolution based on~\cite{dai2017deformableconvolutionalnetworks} has been (unsuccessfully) used for generative tasks and is not identical with our formulation (which is based on ~\cite{DBLP:journals/corr/abs-1811-11168}. In Contrast to ~\cite{spitznagel2026newkindnetworkreview}, our per-tap modulation lets a cell suppress or amplify individual sampling positions rather than only displace them. This allows different heads to specialize to complementary constraint directions, which is critical for spatial reasoning.

\section{Empirical Evaluation}
\vspace{-0.25cm}
\subsection{Solving \textit{Sudoku} on Array Representations}
We evaluate on the held-out test split of the SRM Sudoku dataset~\cite{wewer25srm}, grouping puzzles by difficulty into Easy ($1$--$27$ removed cells), Medium ($28$--$54$), and Hard ($55$--$81$) buckets. Two metrics are reported per bucket: the \textit{Average Solve Rate} (ASR), the fraction of test puzzles whose final NCA state is a valid Sudoku completion of the given clues, and the \textit{Average Cell Precision} (ACP), the fraction of non-clue cells whose predicted digit is consistent with a valid completion. Because high-difficulty puzzles often admit several valid completions, ASR is the primary metric. \\
\Cref{tab:sudoku_array} compares the perception variants at matched cell-state size and network capacity, reporting mean and standard deviation over four training runs each. All variants solve Easy and Medium puzzles near-perfectly, so the evaluation performance is decided on the Hard buckets, where aNCA leads on both metrics while remaining comparable in parameter count. The aNCA performance gain can not be explained by receptive-field size or capacity: i.e. enlarging the fixed kernel beyond $5\times5$ does not increase the performance, $7\times7$ and $9\times9$ variants even perform worse on Hard despite using more parameters then aNCAs. The $9\times9$ variant is also markedly unstable across seeds ($\pm 37.15$ on Hard), whereas the smaller fixed kernels are reproducible.
\begin{table}[h!]
\vspace{-0.25cm}
  \caption{Sudoku on the array representation, reported as mean $\pm$ standard deviation over four training runs per perception. Best Hard-bucket ASR/ACP in bold.}
  \label{tab:sudoku_array}
  \centering
  \scriptsize
  \setlength{\tabcolsep}{2pt}
  \begin{tabular}{lcc|cc|cc|c}
    \toprule
    \multirow{2}{*}{Perception}
      & \multicolumn{2}{c}{Easy [1, 27]}
      & \multicolumn{2}{c}{Medium [28, 54]}
      & \multicolumn{2}{c}{Hard [55, 81]}
      & Param. \\
    \cmidrule(lr){2-3} \cmidrule(lr){4-5} \cmidrule(lr){6-7}
      & ASR & ACP & ASR & ACP & ASR & ACP \\
    \midrule
    NCA w. 3x3 Conv. & 100 $\pm$ 0 & 100 $\pm$ 0 & 86.35 $\pm$ 1.00 & 98.85 $\pm$ 0.13 & 78.23 $\pm$ 1.84 & 98.63 $\pm$ 0.10 & 50k \\
    NCA w. 5x5 Conv. & 100 $\pm$ 0 & 100 $\pm$ 0 & 87.47 $\pm$ 1.65 & 98.93 $\pm$ 0.12 & 75.03 $\pm$ 8.38 & 96.27 $\pm$ 4.05 & 105k \\
    NCA w. 7x7 Conv. & 100 $\pm$ 0 & 100 $\pm$ 0 & 82.70 $\pm$ 1.62 & 98.43 $\pm$ 0.21 & 76.95 $\pm$ 2.07 & 98.43 $\pm$ 0.15 & 189k \\
    NCA w. 9x9 Conv. & 100 $\pm$ 0 & 100 $\pm$ 0 & 71.58 $\pm$ 18.82 & 95.88 $\pm$ 4.41 & 39.80 $\pm$ 37.15 & 84.95 $\pm$ 19.71 & 300k \\
    NCA w. R/C/B Conv. & 100 $\pm$ 0 & 100 $\pm$ 0 & 87.38 $\pm$ 2.64 & 98.90 $\pm$ 0.22 & 78.18 $\pm$ 2.81 & 98.58 $\pm$ 0.19 & 145k \\
    NCA w. Dilated Conv. & 100 $\pm$ 0 & 100 $\pm$ 0 & 86.98 $\pm$ 1.64 & 98.88 $\pm$ 0.17 & 77.23 $\pm$ 10.43 & 98.48 $\pm$ 0.67 & 150k \\
    NCA w. DCNv1 & 99.9 $\pm$ 0.14 & 100 $\pm$ 0 & 75.78 $\pm$ 6.93 & 97.75 $\pm$ 0.7 & 64.55 $\pm$ 10.43 & 97.5 $\pm$ 0.58 & 159k\\
    \hline
    aNCA & 100 $\pm$ 0 & 100 $\pm$ 0 & \textbf{89.50 $\pm$ 4.66} & \textbf{99.08 $\pm$ 0.40} & \textbf{87.78 $\pm$ 6.74} & \textbf{99.18 $\pm$ 0.46} & 165k \\
    \bottomrule
  \end{tabular}
  \vspace{-0.25cm}
\end{table}

\noindent\textbf{Visualizing Learned Spatial Relations.}
\label{sec:visualizing_spatial_relations}
A central appeal of the adaptive perception module is that the predicted per-cell offsets are themselves interpretable: for every Sudoku cell position at every timestep, they allow to reconstruct to which other cells the model learned to attend. We exploit this to ask whether the adaptive perception learns Sudoku's row, column, and $3\times 3$-box constraint graph from data, despite this is never explicitly modeled.
\begin{figure}[h!]
  \centering
  \begin{subfigure}[t]{0.49\linewidth}
    \centering
    \includegraphics[width=\linewidth]{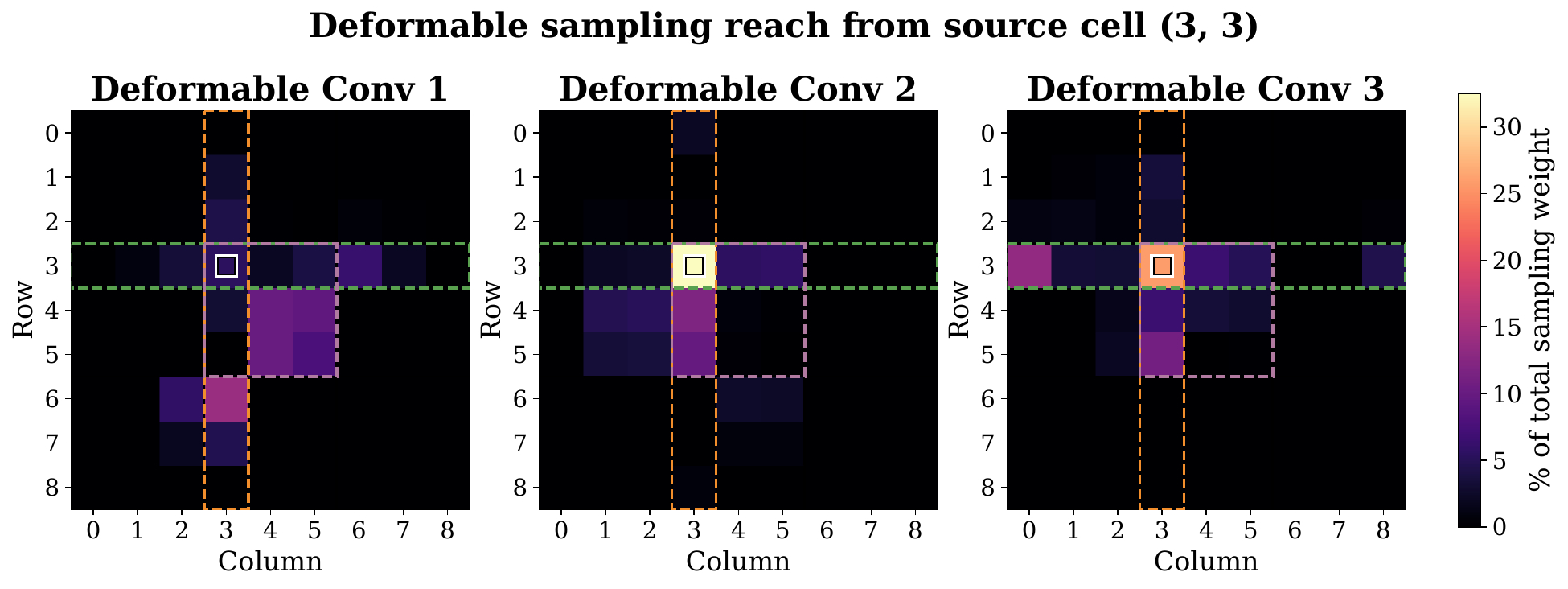}
    \caption{Source cell $(3, 3)$.}
    \label{fig:deform_per_head_3_3}
  \end{subfigure}
  \hfill
  \begin{subfigure}[t]{0.49\linewidth}
    \centering
    \includegraphics[width=\linewidth]{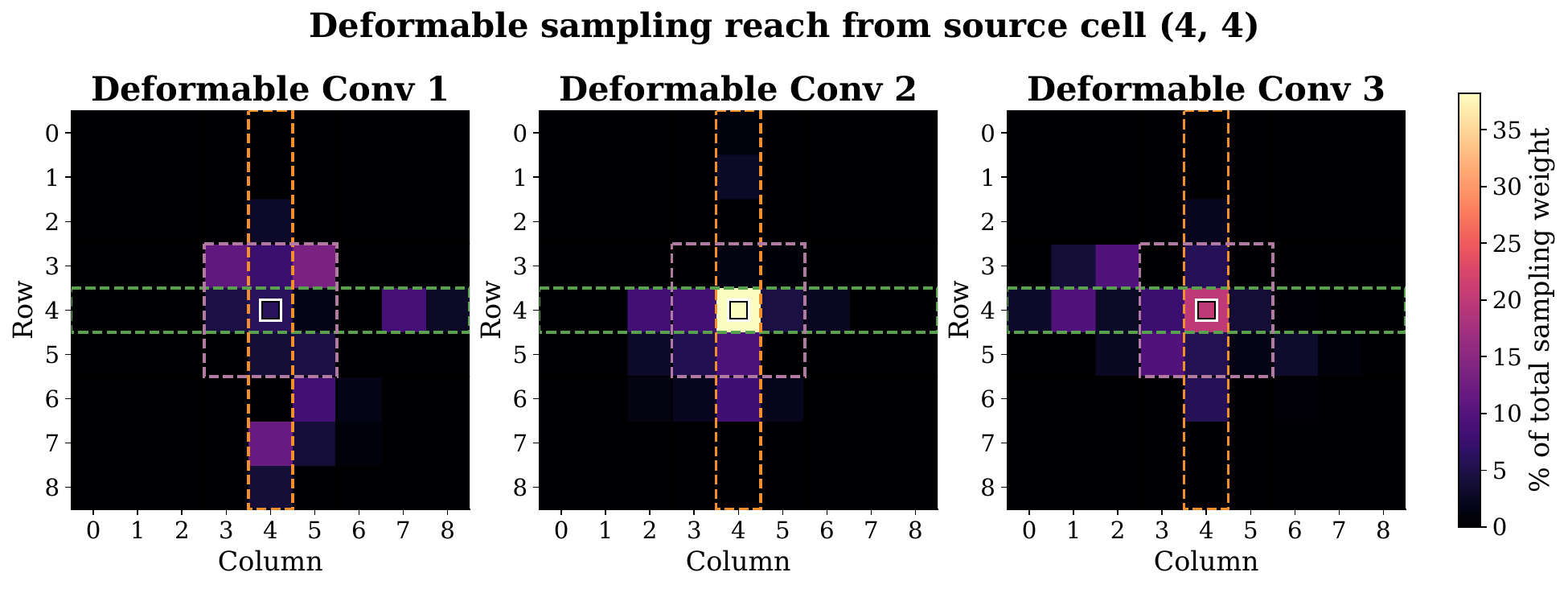}
    \caption{Source cell $(4, 4)$.}
    \label{fig:deform_per_head_4_4}
  \end{subfigure}
  \caption{Learned sampling reach of each deformable perception head from two interior source cells (black-bordered squares at grid positions $(3, 3)$ and $(4, 4)$). Each panel is one head: the heatmap shows the fraction of the head's total sampling weight placed on every cell of the $9\times 9$ board, aggregated over held-out test puzzles at the final NCA step. Dashed overlays mark the source cell's row (green), column (orange), and $3\times 3$ box (purple). The learned heat concentrates inside these overlays and the three heads specialize to complementary constraint directions.}
  \label{fig:deform_per_head}
  \vspace{-0.5cm}
\end{figure}
For each test puzzle we run the trained NCA for the standard rollout horizon and record every head's tap offsets at the final step. Fixing one source cell on the $9\times 9$ board, we then aggregate where the nine taps of each head end up sampling from that cell across held-out testset puzzles. \Cref{fig:deform_per_head} shows the result as one heatmap per head, for two interior source cells ($(3, 3)$ at the intersection of the top-left $3\times 3$ box with the center box, and $(4, 4)$ at the center of the board). 

Dashed rectangles overlay the source cell's row, column, and $3\times 3$ box, so the reader can directly compare the learned hot cells against Sudoku constraints. Across both positions and all heads, the heat concentrates inside those three rectangles. The heads also specialize, with different deformable convolutions preferring the row, column, or box direction, so that together they cover the full constraint graph of the source. Quantitative alignment scores are reported in \cref{tab:alignment}.
\vspace{-0.25cm}
\subsection{Solving \textit{Sudoku} on Images}
\begin{wraptable}{r}{0.5\textwidth}
  \caption{MNIST Sudoku. Results in ASR (Average Solve Rate), reported  as mean $\pm$ standard deviation over four training runs. Best Hard-bucket ASR in bold. Per-bucket ACP is reported in \cref{tab:sudoku_pixel_appendix}.}
  \label{tab:sudoku_pixel}
  \centering
      \scriptsize
  \setlength{\tabcolsep}{2pt}
  \begin{tabular}{lc|c|c|c}
    \toprule
    \multirow{2}{*}{Method}
      & Easy
      & Medium
      & Hard
        \\
      & [1, 27]
      & [28, 54]
      & [55, 81]
      & Param. \\
    \midrule
    Diffusion Model  & 99.4 & 53.6 & 0.8 & 118M \\
    SRM &  99.8 & 75.4 & 51.6 & 118M \\
    \midrule
    &&&& \textit{Codec + NCA} \\
    NCA 3x3 Conv.  & 99.40 $\pm$ 0.14 & 80.90 $\pm$ 0.90 & 71.98 $\pm$ 1.61 & 1.1M + 50k \\
    NCA 5x5 Conv.  & 99.13 $\pm$ 0.50 & 79.97 $\pm$ 1.70 & 69.00 $\pm$ 6.63 & 1.1M + 105k \\
    NCA 7x7 Conv.  & 98.73 $\pm$ 0.10 & 76.55 $\pm$ 1.21 & 72.30 $\pm$ 2.60 & 1.1M + 189k \\
    NCA 9x9 Conv.  & 99.30 $\pm$ 0.23 & 66.73 $\pm$ 16.99 & 35.83 $\pm$ 33.56 & 1.1M + 300k \\
    NCA R/C/B Conv. & 99.20 $\pm$ 0.08 & 80.38 $\pm$ 1.41 & 71.25 $\pm$ 2.61 & 1.1M + 145k \\
    NCA Dilated Conv.  & 99.25 $\pm$ 0.17 & 80.78 $\pm$ 0.95 & 70.95 $\pm$ 8.39 & 1.1M + 150k \\
    NCA DCNv1  & 98.43 $\pm$ 0.54 & 70.15 $\pm$ 6.79 & 60.13 $\pm$ 9.75 & 1.1M + 159k \\
    \hline
    aNCA  & 99.00 $\pm$ 0.37 & \textbf{81.90 $\pm$ 5.50} & \textbf{82.10 $\pm$ 4.89} & 1.1M + 165k \\
    \bottomrule
  \end{tabular}

\vspace{-0.25cm}
\end{wraptable}
We extend the perception comparison to the image-level Sudoku task, in which each cell is rendered as an MNIST-style digit glyph~\cite{wewer25srm}. The NCA operates in the latent board space defined by the frozen encoder--decoder pipeline of \cref{fig:pixel_sudoku_nca}. Held-out splits, difficulty buckets, and metrics are reused, with ASR evaluated with a frozen MNIST classification CNN on the decoded pixel image. We further compare against the two image-space Sudoku solvers released with the dataset~\cite{wewer25srm}: a $118\textnormal{M}$-parameter diffusion model and SRM at the same capacity. Unlike these end-to-end baselines, our pipeline decouples digit recognition from constraint reasoning via the frozen codec. This modularity potentially allows the same reasoner to operate on any object encoding for which a codec maps inputs to the board representation. Codec isolation accuracy is reported in \cref{app:codec_accuracy}.
\Cref{tab:sudoku_pixel} reports the results. aNCA surpass both reference models on every difficulty bucket while using fewer parameters. Among the NCA variants themselves, the ordering matches the array task, with aNCA ahead on Medium and Hard. The parameter column lists the shared codec and the NCA separately. The reported NCAs are the \emph{exact} checkpoints evaluated in \cref{tab:sudoku_array} on the array task. 

\begin{figure}[h!]
\vspace{-0.25cm}
  \centering

  \begin{subfigure}[t]{0.49\linewidth}
    \centering
    \includegraphics[width=\linewidth]{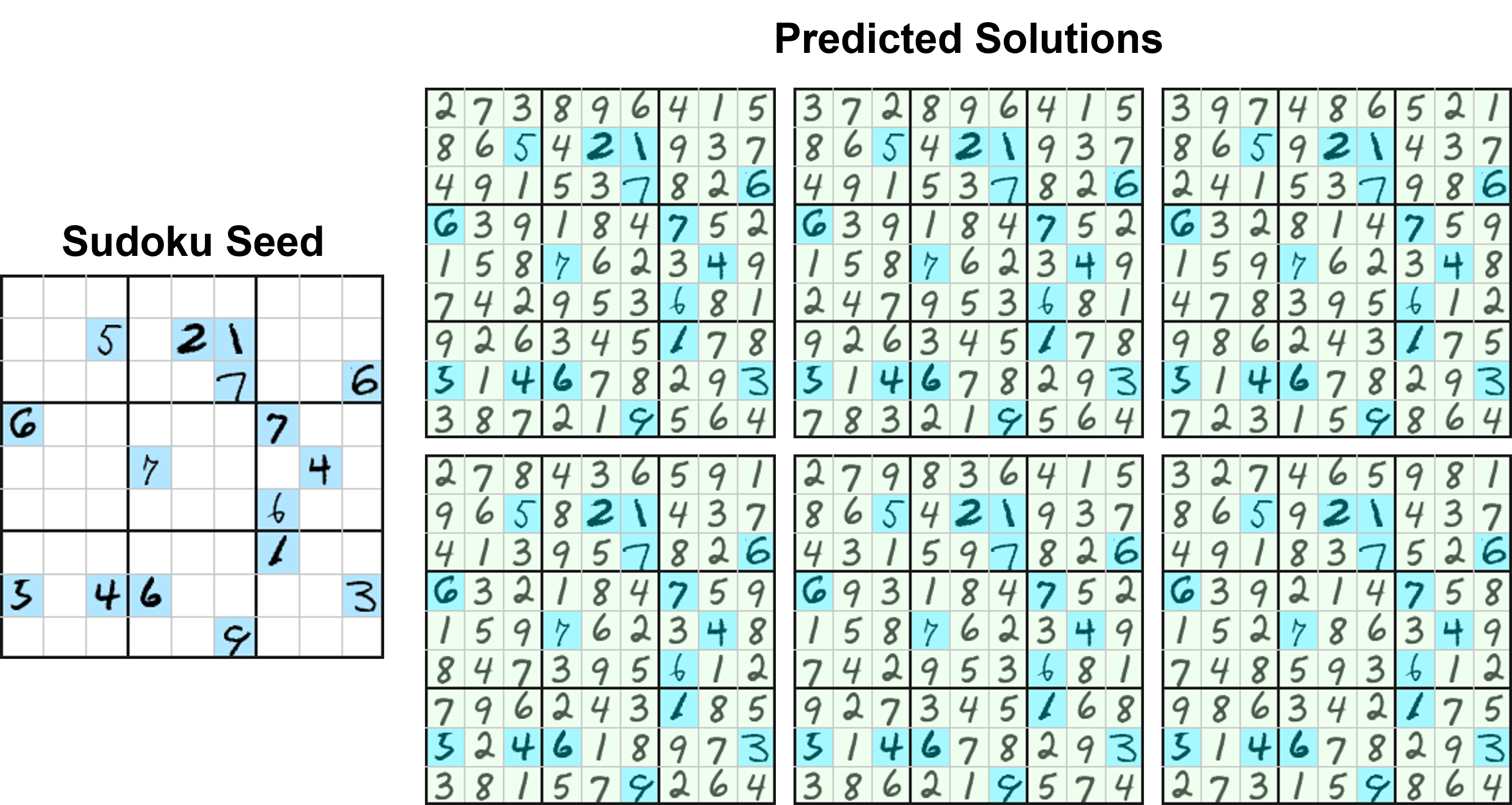}
    \caption{}
    \label{fig:pixel_sudoku_solution_diversity}
  \end{subfigure}
  \hfill
  \begin{subfigure}[t]{0.49\linewidth}
    \centering
    \includegraphics[width=\linewidth]{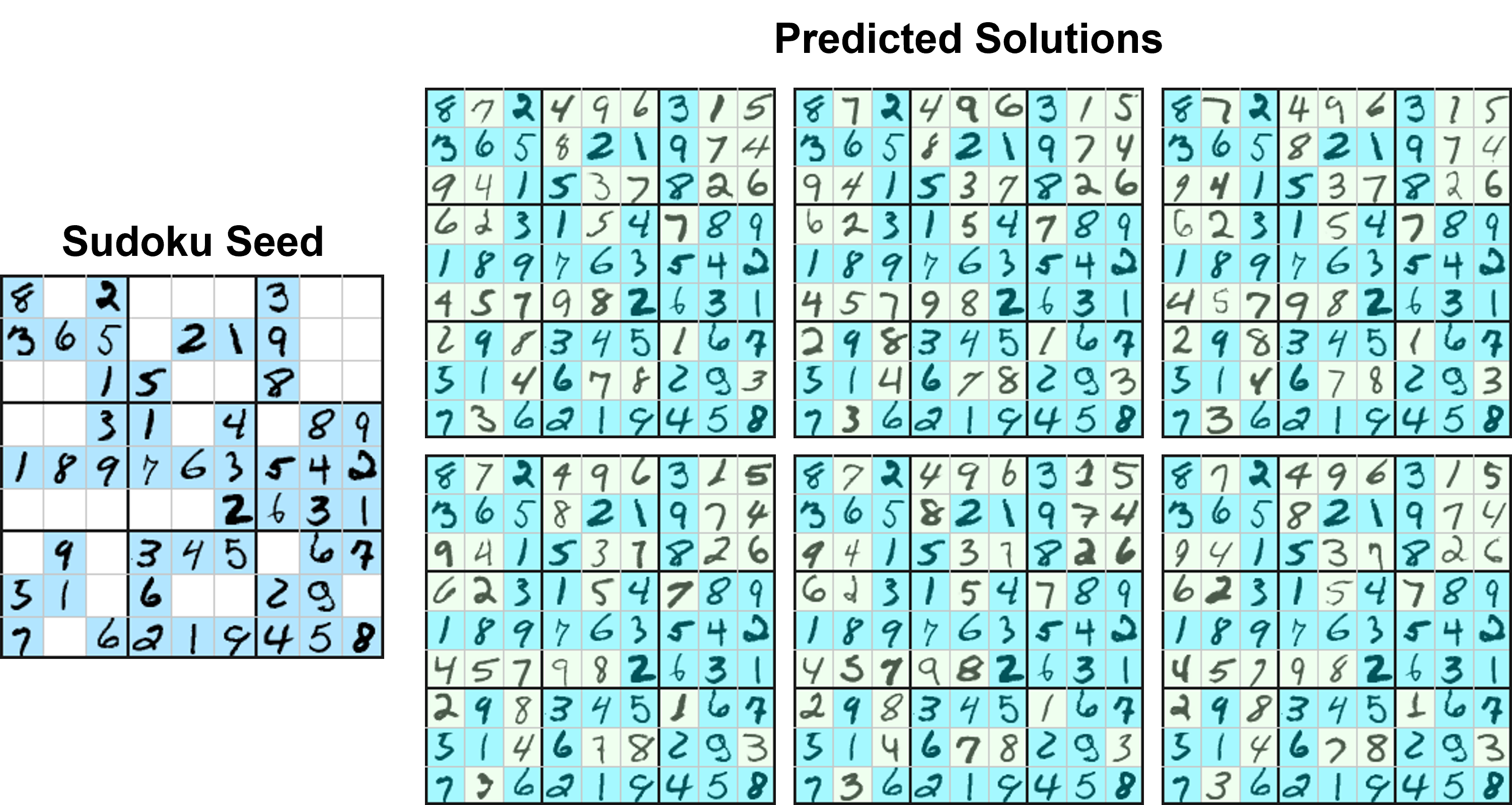}
    \caption{}
    \label{fig:pixel_sudoku_digit_diversity}
  \end{subfigure}
  \caption{Output diversity of the Sudoku codec. (a) For an underconstrained clue pattern with at least $10^6$ valid completions, repeated stochastic NCA rollouts produce multiple distinct valid solutions. (b) For a puzzle with a unique solution, resampling the decoder style latent preserves the solved board while varying the digit style.}
  \label{fig:pixel_sudoku_diversity}
\end{figure}

\noindent\textbf{Output Diversity.}
Highly underconstrained Sudoku puzzles admit many valid completions, so matching the dataset target is neither necessary nor desirable. Sampling the same NCA 1000 times on a puzzle with at least $10^6$ valid completions yields valid completions in 997/1000 rollouts and 303 distinct boards (\cref{fig:pixel_sudoku_solution_diversity}). For a puzzle with a unique solution, resampling produces identical boards but varied digit stroke styles (\cref{fig:pixel_sudoku_digit_diversity}).

\subsection{Solving Maze Puzzles}

\begin{figure}[h]
  \centering
  \vspace{-1cm}
  \begin{subfigure}[t]{0.49\linewidth}
    \centering
    \includegraphics[width=\linewidth]{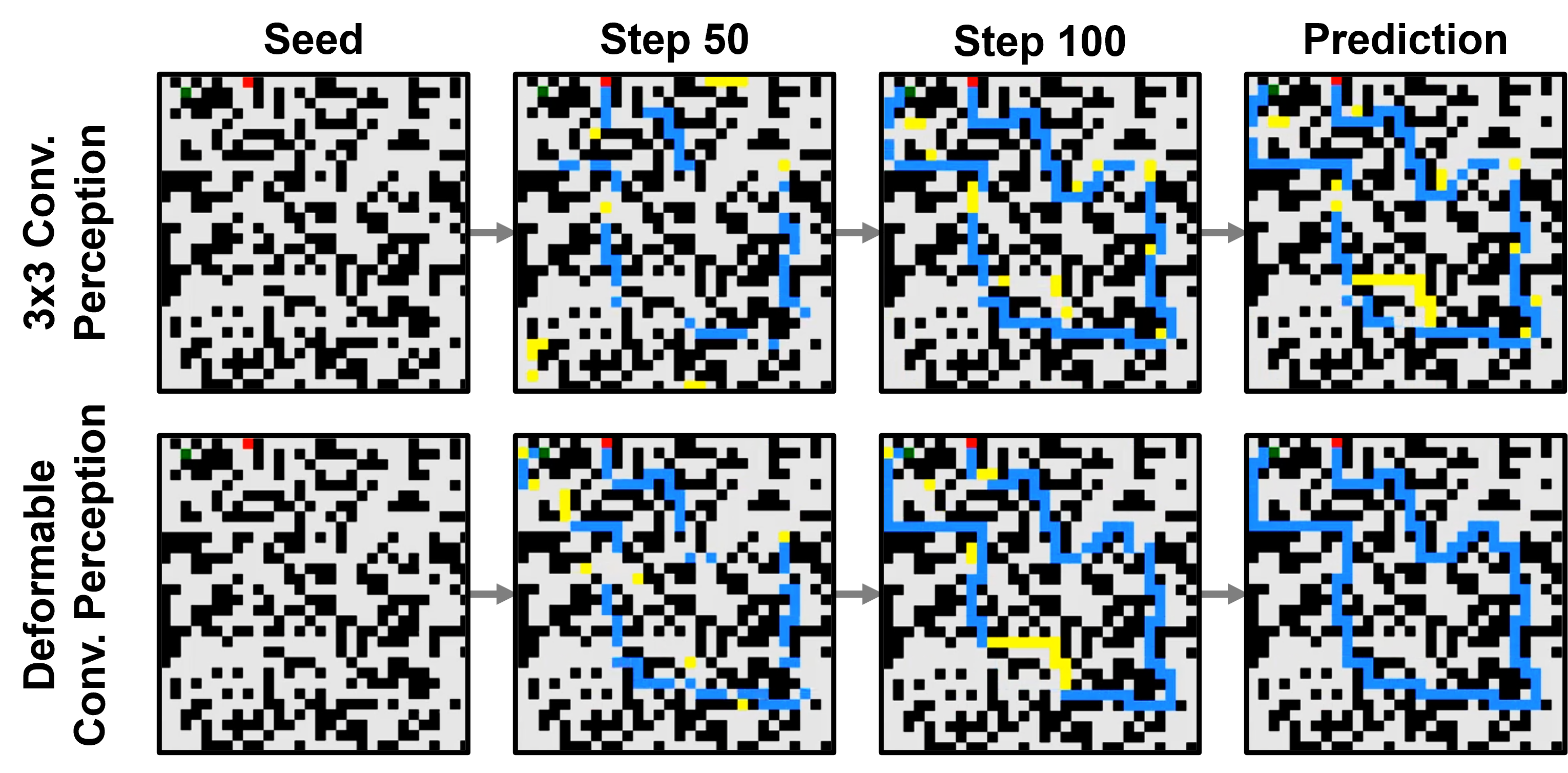}
    \caption{}
    \label{fig:maze_30x30}
  \end{subfigure}
  \hfill
  \begin{subfigure}[t]{0.49\linewidth}
    \centering
    \includegraphics[width=\linewidth]{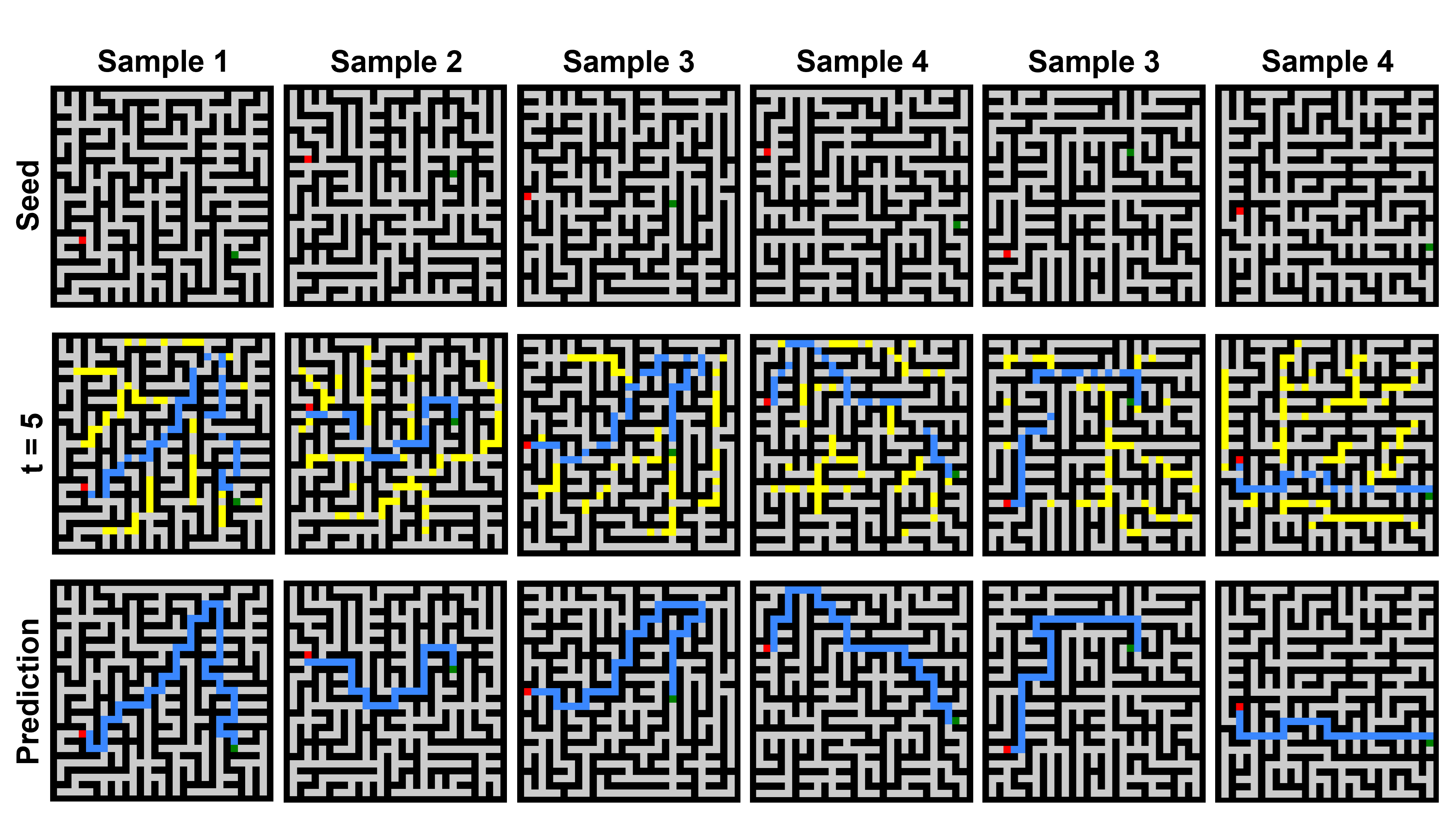}
    \caption{}
    \label{fig:maze_generalization_eval}
  \end{subfigure}

  \caption{(a) Maze solving: the model receives walls (black), start (green), and goal (red). Blue cells indicate correct path predictions matching the target; yellow cells are predicted path pixels that deviate from the target. (b) Generalization to unseen $30\times30$ Prim-style maze structures.}
  \label{fig:secondary_tasks}
\end{figure}

\begin{wraptable}{r}{0.5\textwidth}
\vspace{-0.5cm}
  \caption{Maze solving on $30\times30$ grids, reported as mean $\pm$ standard deviation over five training runs. Valid Path checks connectivity and wall avoidance. Valid Optimal additionally requires shortest-path length.}
  \label{tab:maze}
  \centering
    \scriptsize
  \setlength{\tabcolsep}{1.5pt}
  \resizebox{\linewidth}{!}{%
  \begin{tabular}{lccc|c}
    \toprule
    Method & Source & Valid Path & Valid Optimal & Param. \\
    \midrule
    HRM & \cite{wang2025hierarchicalreasoningmodel} & --$^\star$ & 74.5\% & 27M\\
    ItrSA++ & \cite{kubo2026cvoting} & --$^\star$ & 78.6\% & 3M \\
    TRM-Att & \cite{jolicoeurmartineau2025morerecursivereasoningtiny} & --$^\star$ & 85.3\% & 7M \\
    \midrule
    NCA w. 3x3 Conv. & Ours & 2.54 $\pm$ 0.84 & 0.08 $\pm$ 0.08 & 50k \\
    NCA w. 5x5 Conv. & Ours & 24.78 $\pm$ 1.68 & 10.78 $\pm$ 1.21 & 105k \\
    NCA w. Dilated Conv. & Ours & 74.26 $\pm$ 1.07 & 64.30 $\pm$ 1.32 & 150k \\
    NCA w. R/C/B Conv. & Ours & 93.78 $\pm$ 1.99 & 81.26 $\pm$ 2.01 & 280k \\
    NCA w. DCNv1   & Ours & 86.73 $\pm$ 4.58 & 77.48 $\pm$ 6.42 &  236k \\ 
    \hline
    aNCA & Ours & \textbf{98.56 $\pm$ 0.72} & \textbf{94.18 $\pm$ 1.14} & 256k \\
    \bottomrule
    \multicolumn{5}{l}{ $^\star$ indicates metrics not reported in prior work.}\\
  \end{tabular}
  }
\end{wraptable}
We evaluate on the $30\times30$ mazes described in \cref{sec:spatial_reasoning} and report two graph-level metrics: \emph{Valid Path}, the fraction of predictions that connect start and goal without crossing walls, and \emph{Valid Optimal}, the fraction whose path length matches the shortest-path target. \\
\Cref{tab:maze} shows that purely local perception is insufficient: the $3\times3$ NCA almost never forms a complete path. Enlarging the communication radius closes much of this gap, with dilated perception reaching $64.30\%$ optimality. Adaptive NCAs perform best, producing valid paths for $98.56\%$ of mazes and optimal paths for $95.2\%$, surpassing HRM, ItrSA++ and TRM-Att while using fewer parameters. The R/C/B variant also transfers well despite being designed for Sudoku, but trails deformable perception, suggesting that content-adaptive routing is more useful than a fixed constraint geometry on mazes. Same-resolution transfer to a procedurally distinct maze distribution is reported in \cref{app:maze_results}.

\vspace{-0.25cm}
\subsection{Learning to Adhere Global Constraints}
\begin{wraptable}{r}{0.5\textwidth}
\vspace{-0.5cm}
  \caption{Global color balance. NCA results are mean $\pm$ standard deviation across independent training runs. Imbalance measures absolute color-count error. Accuracy measures exact satisfaction of the $50/50$ ratio. }
  \label{tab:color_imbalance}
  \centering
\scriptsize
  \setlength{\tabcolsep}{1.5pt}
  \resizebox{\linewidth}{!}{%
  \begin{tabular}{lccc|c}
    \toprule
    Perception & Source & Imbalance $\downarrow$ & Accuracy $\uparrow$ & Param. \\
    \midrule
    Diffusion & \cite{wewer25srm} & 1.27 & 25.0\% & 19.7M\\
    SRM & \cite{wewer25srm} & 0.53 & 51.8\% & 19.7M\\
    \midrule
    NCA w. 3x3 Conv. & Ours & 1.13$\pm$0.17 & 26.0\%$\pm$5\% & 48k \\
    NCA w. 5x5 Conv. & Ours & \textbf{0.07$\pm$0.06} & 95.3\%$\pm$2\% & 100k \\
    NCA w. DCNv1   & Ours & 1.61$\pm$0.59 & 18.7\%$\pm$11\%  & 147k\\ 
    \hline
    aNCA & Ours & 0.1$\pm$0.08 & \textbf{96.8\%$\pm$3\%} & 152k\\
    \bottomrule
    \vspace{-0.5cm}
  \end{tabular}
  }
\end{wraptable}
The color-balance task tests whether the NCA can satisfy a constraint defined over the entire grid, without local target labels or a conditioning image. We evaluate the output by its absolute color-count error and exact-ratio accuracy, reporting mean and standard deviation across independent training runs in \cref{tab:color_imbalance}.
\Cref{tab:color_imbalance} shows that aNCA achieves the highest mean exact-ratio accuracy among the compared models while using substantially fewer parameters than the diffusion and SRM references. The $5\times5$ NCA remains competitive and achieves the lowest mean count error. These metrics capture different aspects of performance: exact accuracy measures how often the global constraint is satisfied, whereas mean count error measures the average deviation from the $50/50$ balance.

These results show that adaptive perception also helps for a spatially uniform objective. In \cref{tab:color_imbalance}, aNCA with modulated DCNv2 clearly improves exact accuracy over offset-only DCNv1, while gains over the $5\times5$ baseline are smaller and within run-to-run variability (also see \cref{fig:color_balance} for qualitative results).

\section{Conclusion}

We introduced \textit{Adaptive Neural Cellular Automata} (aNCA), which implement a conceptually novel perception approach to enable NCAs to learn reasoning capabilities. On spatially structured tasks, Sudoku, image-based Sudoku, and maze solving, aNCA sets new state-of-the-art results while using a fraction of the parameters of prior methods. The learned perceptions are directly inspectable, revealing that the model learns task-relevant spatial relations (e.g.\ Sudoku rows, columns, and boxes) without explicit encoding in the architecture. 

\noindent\textbf{Limitations.} \label{sec:limitations}
While aNCA achieves strong results on spatially structured tasks. However, its adaptive perception comes at higher per-step computational cost than fixed-convolution baselines. Detailed resource measurements are provided in~\cref{app:sudoku_resource_usage}. The current architecture targets regular 2D grids. Adapting its learned perception mechanism to irregular graphs, 3D scenes, and variable-resolution inputs is a promising direction for future work.

\subsection*{Funding Acknowledgement}
The authors acknowledge the financial support by the German Federal 
Ministry of Research, Technology and Space (BMFTR) in the program “Forschung an Fachhochschulen in Kooperation mit Unternehmen (FH-Kooperativ)” within the joint project "KI-Bohrer" under 
grant 13FH525KX1.

\subsection*{AI use statement}

In this work, we used generative AI tools as a coding assistant and for language editing, to revise and shorten passages of the manuscript that we had drafted ourselves. Generative AI tools were not used to generate research ideas, to design the proposed architecture, to produce or interpret experimental results, or to retrieve or summarize related work. All code was reviewed and tested by the authors, and every reported number was produced by running that code on the checkpoints described in \cref{app:implementation}. We have reviewed all AI-assisted text and take full responsibility for the final content of this work, including all text, claims and artifacts.

\subsection*{Ethics statement}

This work studies architectures for 2D spatial reasoning on synthetic and publicly available benchmarks. It involves no human subjects, no personally identifiable information, and no user data. All datasets used are public and were employed within their intended scope: the Sudoku and MNIST-Sudoku data of~\cite{wewer25srm} and the maze dataset of~\cite{wang2025hierarchicalreasoningmodel}. We release no new dataset. The proposed models are small and were trained on a single consumer GPU, so the computational footprint of this work is modest. We see no direct pathway from puzzle solving to harmful application, although improved spatial reasoning is a general capability whose downstream uses we cannot fully anticipate. The authors declare no conflicts of interest.

\subsection*{Reproducibility statement}

All datasets and benchmarks used in this work are publicly available and are cited in \cref{sec:spatial_reasoning}. Model and training hyperparameters for each task are given in \cref{app:hyperparameters_sudoku,app:hyperparameters_maze,app:hyperparameters_color_balance}, the exact training objectives in \cref{app:loss_sudoku,app:loss_maze,app:loss_color_balance}, and the encoder-decoder codec used for MNIST Sudoku in \cref{app:codec,app:codec_loss}, together with an isolation evaluation of its accuracy in \cref{app:codec_accuracy}. Computational cost and hardware are reported in \cref{app:sudoku_resource_usage} and \cref{app:implementation}. All reported results are means and standard deviations over multiple independent training runs rather than single runs, with per-seed behavior discussed in \cref{app:sudoku_results}. The full source code, including the configuration files reproducing every reported run, will be released upon publication as a plugin for the \textit{NCATorch} framework of~\cite{spitznagel2026newkindnetworkreview}.



\bibliography{iclr2027_conference}
\bibliographystyle{unsrt}
\newpage
\appendix
\section{Technical appendices and supplementary material}


\noindent\textbf{Contents}
\etocsettocstyle{}{}
\etocsettocdepth{subsubsection}
\localtableofcontents

\subsection{Video Material}
\label{app:videos}
\url{
https://youtu.be/WF0Bm6Ie7WA
}
\begin{itemize}
    \item \href{https://youtu.be/WF0Bm6Ie7WA}{0:00 Sudoku: aNCA Solving (Batch)}
    \item \href{https://youtu.be/WF0Bm6Ie7WA?si=xTHSlpgJ4kyRxKIh&t=16}{0:16 Sudoku: aNCA Solving (Hard)}
    \item \href{https://youtu.be/WF0Bm6Ie7WA?si=xTHSlpgJ4kyRxKIh&t=47}{0:47 Sudoku: aNCA Output Diversity}
    \item \href{https://youtu.be/WF0Bm6Ie7WA?si=27fbd0FSeP_w-Yez&t=70}{1:10 Sudoku MNIST: aNCA Solving (Batch)}
    \item \href{https://youtu.be/WF0Bm6Ie7WA?si=r54PXpzD0h2ot815&t=107}{1:47 Sudoku MNIST: aNCA Solving (Hard)}
    \item \href{https://youtu.be/WF0Bm6Ie7WA?si=46ALbtc2JKkyFQzq&t=130}{2:10 Sudoku MNIST: aNCA Output Diversity (Solutions)}
    \item \href{https://youtu.be/WF0Bm6Ie7WA?si=0JhQjRuaDrgDtw_0&t=145}{2:25 Sudoku MNIST: aNCA Output Diversity (Digit Style)}
    \item \href{https://youtu.be/WF0Bm6Ie7WA?si=LApurAzYp7KtES9f&t=162}{2:42 Sudoku MNIST: aNCA learned spatial relations}
    \item \href{https://youtu.be/WF0Bm6Ie7WA?si=q9I-SKgKednKf5id&t=294}{4:54 Maze 30x30: NCA w. 3x3 Conv.}
    \item \href{https://youtu.be/WF0Bm6Ie7WA?si=sSnvEz9HIUg_SoTA&t=307}{5:07 Maze 30x30: aNCA}
    \item \href{https://youtu.be/WF0Bm6Ie7WA?si=2XkOw4aZa-xsgFSd&t=320}{5:20 Maze DFS 20x20 (OOD): aNCA}
    \item \href{https://youtu.be/WF0Bm6Ie7WA?si=xzshJngsfCAt4zwf&t=337}{5:37 Maze Prim 40x40 (OOD): aNCA}
    \item \href{https://youtu.be/WF0Bm6Ie7WA?si=J-9AyRW9Bp_K4G4L&t=345}{5:45 Color Balance Task Perception Comparison}
\end{itemize}

\subsection{Implementation Details}
\label{app:implementation}
All models are implemented in PyTorch. Deformable convolutions use \texttt{torchvision.ops.deform\_conv2d}, the standard torchvision DCNv2 implementation. All experiments were run on a single NVIDIA RTX 4090. To provide reproducibility, the code will be released upon publication, as a plugin for the \textit{NCATorch} framework of~\cite{spitznagel2026newkindnetworkreview}.

\subsubsection{Difference between DCNv1 and DCNv2}
\label{app:dcn_comparison}
Both variants predict $2k^2$ offsets from the current cell state, yielding two displacement coordinates for each of the $k^2$ kernel taps. DCNv1 then samples only at the displaced positions:
\begin{equation}
  \mathcal{P}_\phi^{\mathrm{DCNv1}}(s,\mathbf{p})
  = \sum_{n=1}^{k^2} w_n \cdot
    s\!\left(\mathbf{p}+\mathbf{p}_n+\Delta\mathbf{p}_n(\mathbf{p})\right).
\end{equation}
DCNv2 additionally predicts one content-dependent modulation value per tap, requiring a further $k^2$ outputs, and weights every sampled feature before aggregation:
\begin{equation}
  \mathcal{P}_\phi^{\mathrm{DCNv2}}(s,\mathbf{p})
  = \sum_{n=1}^{k^2} w_n \cdot m_n(\mathbf{p}) \cdot
    s\!\left(\mathbf{p}+\mathbf{p}_n+\Delta\mathbf{p}_n(\mathbf{p})\right).
\end{equation}
Thus, DCNv1 adapts only \emph{where} each tap samples, whereas DCNv2 also adapts \emph{how strongly} each sampled location contributes. In our implementation, $m_n(\mathbf{p})=2\,\sigma(a_n(\mathbf{p}))\in[0,2]$, where $a_n$ is produced by a separate modulation convolution. The offset and modulation predictors are zero-initialized, so the module initially behaves like a regular convolution: offsets are zero and modulation values are one.

\clearpage
\subsection{Sudoku}
\label{app:sudoku}

\subsubsection{Model and Training Hyperparameters}
\label{app:hyperparameters_sudoku}

\begin{table}[h]
  \centering
  \caption{Hyperparameters shared across all NCA variants for the Sudoku task. Perception-specific settings (kernel size, dilation, number of deformable heads) vary per variant and are described in \cref{sec:methods}.}
  \label{tab:hyperparameters_sudoku}
  \setlength{\tabcolsep}{6pt}
  \begin{tabular}{llr}
    \toprule
    Category & Hyperparameter & Value \\
    \midrule
    \multirow{2}{*}{NCA state}
      & Cell channels & 24 \\
      & Positional embeddings & yes \\
    \midrule
    \multirow{1}{*}{Update network}
      & Architecture & MLP ($1{\times}1$ Conv), 1 hidden layer, 128 channels \\
    \midrule
    \multirow{6}{*}{Optimization}
      & Optimizer & AdamW \\
      & Learning rate & $4\times10^{-4}$ \\
      & LR schedule & WSD (warmup--stable--decay) \\
      & Warmup / stable / decay & 10\% / 80\% / 10\% \\
      & Final LR & 0 \\
      & Gradient clip norm & 1.0 \\
    \midrule
    \multirow{3}{*}{Training}
      & Total steps & 100\,000 \\
      & Batch size & 128 \\
      & Rollout steps $T$ & $\mathcal{U}[18,\,24]$ \\
    \midrule
    \multirow{2}{*}{Evaluation}
      & Inference rollout steps & 1\,500 \\
      & Digit prediction & $\operatorname{argmax}$ over 9 channels (no post-processing) \\
    \midrule
    \multirow{2}{*}{Sample pool}
      & Pool size & 512 \\
      & Pool fraction & 0.5 \\
    \bottomrule
  \end{tabular}
\end{table}

\subsubsection{Training Loss}
\label{app:loss_sudoku}

Let $p_{i,d} = \mathrm{softmax}(\hat{s}_i)_d$ be the predicted probability of digit $d \in \{1,\ldots,9\}$ at cell $i$. A valid Sudoku completion requires each digit to appear exactly once in every row, column, and $3\times3$ box. Denoting the set of all 27 such units as $\mathcal{U}$, we penalize squared deviations from the per-unit digit count of one:
\begin{equation}
  \mathcal{L}_\text{constraint} = \frac{1}{|\mathcal{U}|} \sum_{U \in \mathcal{U}} \sum_{d=1}^{9} \left(\sum_{i \in U} p_{i,d} - 1\right)^2.
\end{equation}
An additional cross-entropy term enforces consistency with the given clue cells $\mathcal{C}$ whose correct digit is $y_i$:
\begin{equation}
  \mathcal{L}_\text{given} = -\frac{1}{|\mathcal{C}|}\sum_{i \in \mathcal{C}} \log p_{i,\,y_i}.
\end{equation}
The total loss is $\mathcal{L} = \mathcal{L}_\text{constraint} + \mathcal{L}_\text{given}$.
\clearpage
\subsubsection{Computational Cost}
\label{app:sudoku_resource_usage}

\begin{table}[h]
  \centering
  \caption{Resource usage of the Sudoku NCA variants. Training VRAM is the peak CUDA memory allocated during a representative training step (batch size $128$, rollout length $24$). Inference VRAM and runtime are measured at batch size $1$ with a rollout of $1500$ steps on an NVIDIA RTX 4090. Inference time is averaged per puzzle over repeated runs after warm-up.}
  \label{tab:sudoku_resource_usage}
  \begin{tabular}{lrrrr}
    \toprule
    Variant & Params & Train VRAM (MB) & Infer VRAM (MB) & Infer ms / puzzle \\
    \midrule
    NCA w. 3x3 Conv.        & 50k  & 303.1  & 0.5  & 159.64  \\
    NCA w. 5x5 Conv.        & 105k & 311.9  & 0.9 & 166.38  \\
    NCA w. R/C/B Conv.      & 145k & 610.3  & 1.0 & 251.04  \\
    NCA w. Dilated Perc.    & 150k & 480.8 & 1.2 & 238.28  \\
    aNCA                    & 165k & 1139.6    & 17.7 & 832.75 \\
    \bottomrule
  \end{tabular}
\end{table}

\subsubsection{Additional Results: Sudoku}
\label{app:sudoku_results}
This section supplements the main array-Sudoku evaluation with further analysis of aNCA's adaptive perception and reproducibility. We visualize the sampling-distance distributions of the deformable heads, quantify their alignment with Sudoku's row, column, and box constraints, and report the variation across independent training seeds.

\begin{figure}[h!]
  \centering

  \begin{subfigure}[t]{0.45\linewidth}
    \centering
    \includegraphics[width=\linewidth]{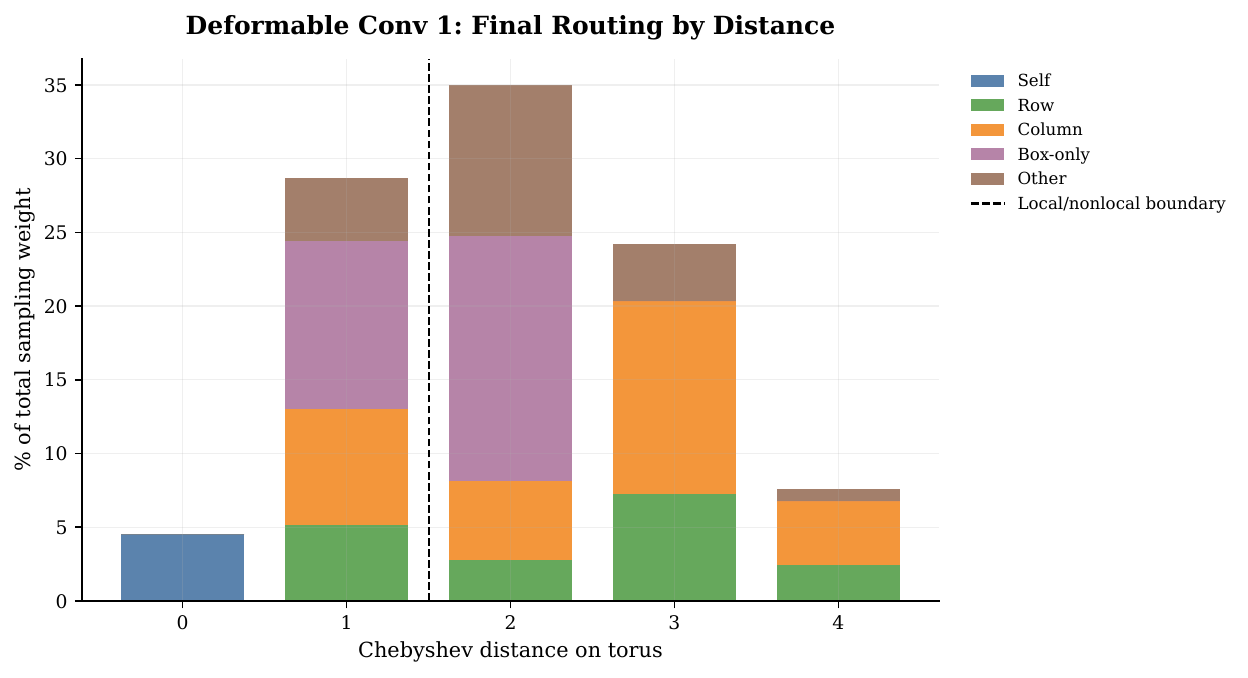}
    \caption{A}
    \label{fig:deform_conv_1}
  \end{subfigure}
  \begin{subfigure}[t]{0.45\linewidth}
    \centering
    \includegraphics[width=\linewidth]{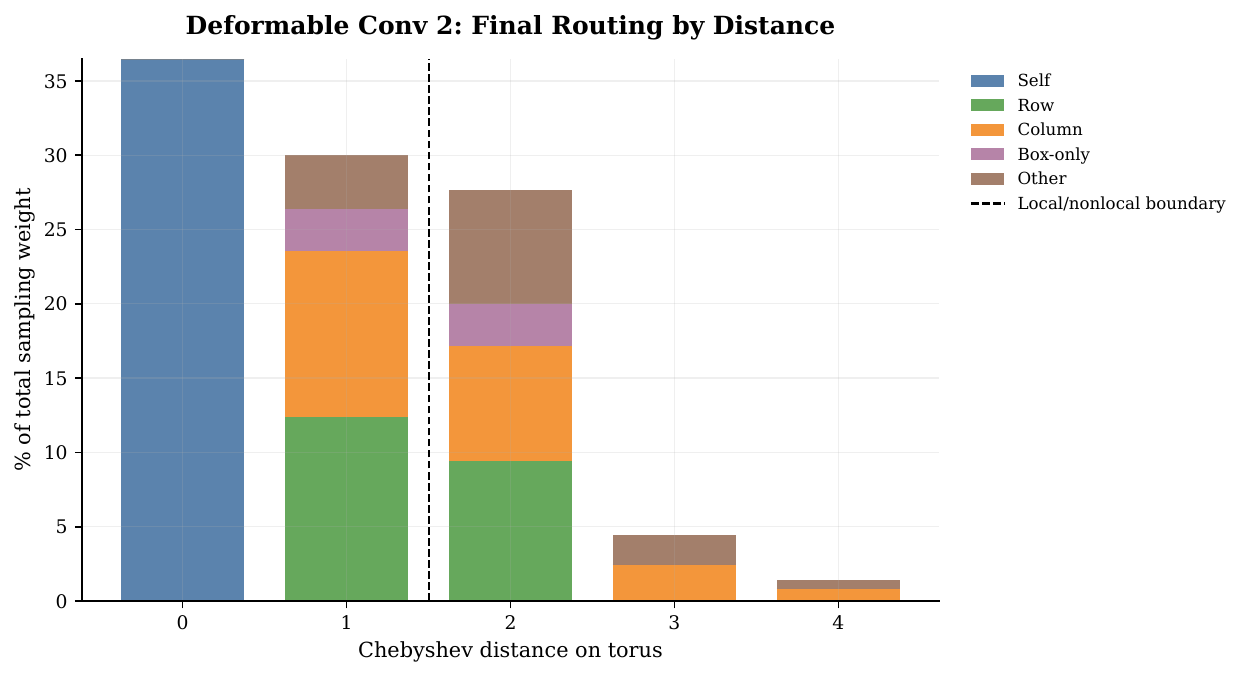}
    \caption{B}
    \label{fig:deform_conv_2}
  \end{subfigure}
\begin{subfigure}[t]{0.45\linewidth}
    \centering
    \includegraphics[width=\linewidth]{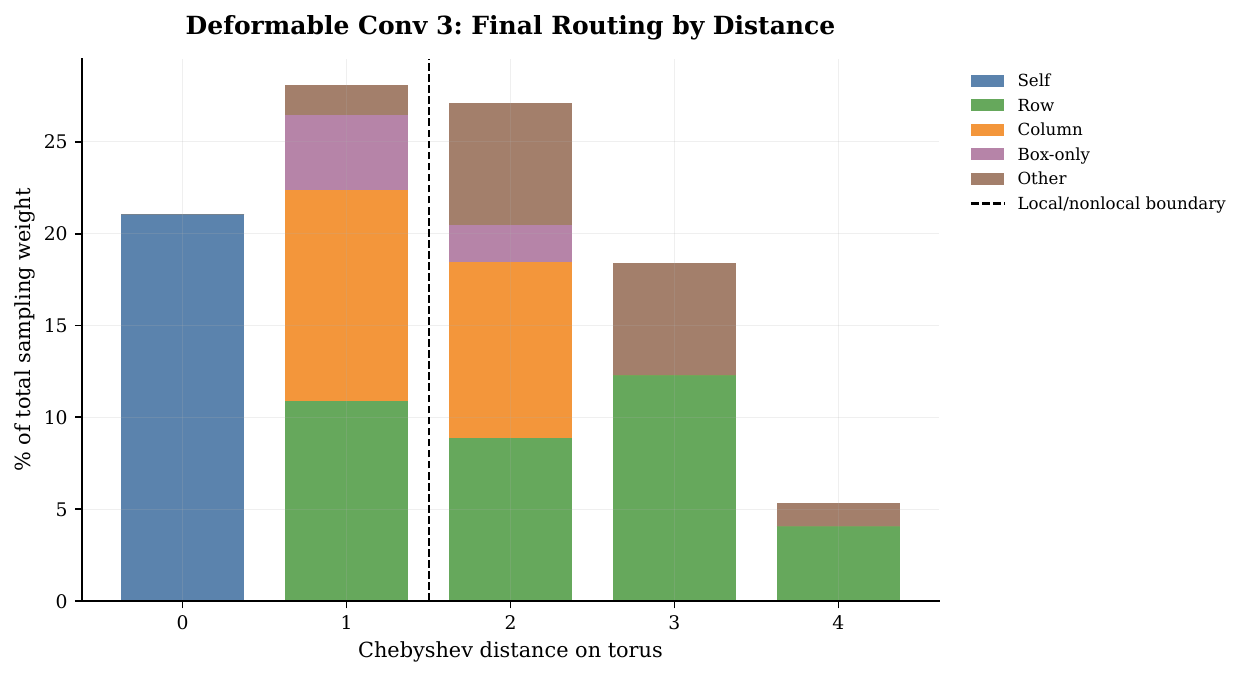}
    \caption{C}
    \label{fig:deform_conv_3}
  \end{subfigure}

  \caption{Learned sampling distance distributions of the three deformable perception heads (A, B, C) at the final NCA step, aggregated over held-out test puzzles. Each plot shows how far each kernel tap reaches from the source cell, revealing that the heads develop complementary routing strategies: some taps remain local while others extend to capture row, column, or box constraints.}
\label{fig:deform_routing_distance}
\end{figure}
\newpage

\noindent\textbf{Quantitative alignment scores.}
\Cref{tab:alignment} reports, for each deformable head, the mean fraction of total sampling weight landing inside the source cell's row, column, $3\times3$ box, and their union (Sudoku-relevant), averaged over all 81 source cells and 10\,000 held-out test puzzles at the final rollout step. The random baseline is the expected fraction under uniform sampling. All three heads concentrate $81$--$87\%$ of their weight on Sudoku-relevant cells, $3.1$--$3.4\times$ above the $25.9\%$ random baseline.

\begin{table}[h]
  \centering
  \caption{Per-head constraint alignment at the final NCA step, averaged over all 81 source cells and 10\,000 test puzzles. Union\% is the fraction of weight on the source cell's row $\cup$ column $\cup$ box. Random baseline is the expected Union\% under uniform sampling ($25.9\%$).}
  \label{tab:alignment}
  \setlength{\tabcolsep}{6pt}
  \begin{tabular}{lrrrrr}
    \toprule
    Head & Row\% & Col\% & Box\% & Union\% & Enrichment \\
    \midrule
    Head 1 & 20.7 & 32.6 & 47.5 & 81.0 & 3.13$\times$ \\
    Head 2 & 61.1 & 59.6 & 69.6 & 86.9 & 3.35$\times$ \\
    Head 3 & 53.0 & 45.8 & 54.3 & 83.5 & 3.22$\times$ \\
    \midrule
    Random baseline & -- & -- & -- & 25.9 & 1.00$\times$ \\
    \bottomrule
  \end{tabular}
\end{table}


\clearpage
\subsection{MNIST Sudoku}
\label{app:mnist_sudoku}

The NCA variants evaluated on MNIST Sudoku are the \emph{exact} checkpoints trained for the array task (\cref{app:sudoku}); no additional NCA training is performed. The pixel-space pipeline wraps these frozen NCA checkpoints with a separately trained encoder--decoder codec, described below.

\subsubsection{Codec Hyperparameters}
\label{app:codec}

\begin{table}[h]
  \centering
  \caption{Architecture and training hyperparameters for the Sudoku codec. Encoder and decoder are trained independently with AdamW.}
  \label{tab:hyperparameters_codec}
  \setlength{\tabcolsep}{6pt}
  \begin{tabular}{llr}
    \toprule
    Category & Hyperparameter & Value \\
    \midrule
    \multirow{2}{*}{Encoder $E_\psi$}
      & Base channels & 32 \\
      & Hidden dim & 128 \\
    \midrule
    \multirow{3}{*}{Decoder $D_\omega$}
      & Base channels & 64 \\
      & Hidden dim & 128 \\
      & Style latent dim & 8 \\
    \midrule
    \multirow{4}{*}{Optimization}
      & Optimizer & AdamW \\
      & Learning rate & $1\times10^{-3}$ \\
      & LR schedule & Warmup--cosine (1000 steps) \\
      & Total steps & 100\,000 \\
    \midrule
    Training & Batch size & 32 \\
    \bottomrule
  \end{tabular}
\end{table}
\clearpage
\subsubsection{Codec Training Loss}
\label{app:codec_loss}

The decoder $D_\omega$ is trained in isolation with a composite loss. Let $\hat{x} = D_\omega(b, z)$ be the rendered image for board $b$ with style latent $z \sim q_\omega(z \mid b)$, $x^*$ the target pixel image, and $M_\text{pix}$ a pixel-space binary mask over occupied (non-blank) cells. The total loss is
\begin{equation}
  \mathcal{L}_\text{dec} = w_\text{KL}\,\mathcal{L}_\text{KL}
  + w_\text{pix}\,\mathcal{L}_\text{pix}
  + w_\text{feat}\,\mathcal{L}_\text{feat}
  + w_\text{cls}\,\mathcal{L}_\text{cls}
  + w_\text{TV}\,\mathcal{L}_\text{TV}
  + w_\text{ink}\,\mathcal{L}_\text{ink}
  + w_\text{bin}\,\mathcal{L}_\text{bin}
\end{equation}
with weights $(w_\text{KL}, w_\text{pix}, w_\text{feat}, w_\text{cls}, w_\text{TV}, w_\text{ink}, w_\text{bin}) = (0.001, 1.0, 0.5, 1.0, 0.01, 0.0025, 0.01)$.

\noindent\textbf{KL divergence.} $\mathcal{L}_\text{KL} = D_\text{KL}(q_\omega(z\mid b)\,\|\,\mathcal{N}(0,I))$, the standard VAE term regularising the style posterior.

\noindent\textbf{Pixel L1.} $\mathcal{L}_\text{pix} = {\sum_i |\hat{x}_i - x^*_i|\cdot M_{\text{pix},i}}\,/\,{\sum_i M_{\text{pix},i}}$, a masked per-pixel reconstruction loss.

\noindent\textbf{Feature L1.} Let $\phi(\cdot)$ denote the pooled intermediate features of the frozen digit classifier $E_\psi$, and let $\hat{c}, c^*$ be the rendered and target cell crops. $\mathcal{L}_\text{feat} = \|\phi(\hat{c}) - \phi(c^*)\|_1$ is a perceptual term encouraging feature-level similarity to the target digit appearance.

\noindent\textbf{Classifier CE.} $\mathcal{L}_\text{cls} = \mathrm{CE}(E_\psi(\hat{c}),\, d^*)$, cross-entropy under the frozen classifier, directly rewarding legible digit rendering.

\noindent\textbf{Total variation.} Let $\mathrm{ink} = (1 - \hat{x})\cdot M_\text{pix}$. Then $\mathcal{L}_\text{TV} = \operatorname{mean}(|\Delta_h\,\mathrm{ink}| + |\Delta_w\,\mathrm{ink}|)$ encourages smooth, connected strokes.

\noindent\textbf{Ink sparsity.} $\mathcal{L}_\text{ink} = {\sum_i (1-\hat{x}_i)\,M_{\text{pix},i}}\,/\,{\sum_i M_{\text{pix},i}}$ penalises dark pixels outside digit regions.

\noindent\textbf{Binary commitment.} $\mathcal{L}_\text{bin} = {\sum_i \hat{x}_i(1-\hat{x}_i)\,M_{\text{pix},i}}\,/\,{\sum_i M_{\text{pix},i}}$ penalises intermediate grey values, pushing pixels toward binary black/white.\\

The encoder $E_\psi$ is trained separately with a masked cross-entropy loss on the given (clue) cells $\mathcal{C}$:
\begin{equation}
  \mathcal{L}_\text{enc} = -\frac{1}{|\mathcal{C}|}\sum_{i\in\mathcal{C}} \log p_{i,\,y_i},
\end{equation}
where $p_{i,d}$ is the predicted probability of digit $d$ at cell $i$ and $y_i$ is the ground-truth digit.

\subsubsection{Codec Accuracy}
\label{app:codec_accuracy}
We evaluate the encoder $E_\psi$ and decoder $D_\omega$ in isolation on 10\,000 test puzzles across the full difficulty range to quantify how much of the pipeline error originates from the codec versus the NCA.

\noindent\textbf{Encoder.} Running $E_\psi$ on the input pixel Sudoku and taking the argmax over digit logits yields $99.38\%$ per-cell accuracy on the given (non-blank) cells. The perfect-board rate (the fraction of boards where every given cell is correctly recognized) is $74.58\%$, meaning roughly one in four boards contains at least one misrecognised clue cell. Despite these imperfect inputs, aNCA maintains strong downstream solving performance (\cref{tab:sudoku_pixel_appendix}) without additional training, highlighting its ability to operate within noisy encoder conditioning.

\noindent\textbf{Decoder round-trip.} We pass the ground-truth one-hot board through $D_\omega$, decode the resulting pixel image with $E_\psi$, and compare the re-encoded digits to the ground truth. Per-cell accuracy is $100\%$ and the perfect round-trip rate is $100\%$: the decoder renders digits reliably enough that the encoder always recovers the correct symbol. Decoder errors therefore contribute zero to the pipeline failure rate. ASR is evaluated against ground-truth clue digits, so encoder misrecognitions affect NCA conditioning but not the validity metric. 
\newpage

\subsubsection{Additional Results: MNIST Sudoku}
\label{app:mnist_sudoku_results}
This section supplements the main MNIST Sudoku evaluation with complete per-bucket results. In addition to ASR, we report ACP for every NCA perception variant on the Easy, Medium, and Hard splits, using the same frozen codec and array-trained NCA checkpoints as in the main evaluation.

\begin{table}[h!]
  \caption{Sudoku Images}
  \label{tab:sudoku_pixel_appendix}
  \centering
  \scriptsize    
  \setlength{\tabcolsep}{2pt}
  \begin{tabular}{lccc|cc|cc|c}
  
    \toprule
    \multirow{2}{*}{Method}
      & Source
      & \multicolumn{2}{c}{Easy [1, 27]} 
      & \multicolumn{2}{c}{Medium [28, 54]} 
      & \multicolumn{2}{c}{Hard [55, 81]} 
      & Param. \\
    \cmidrule(lr){3-4} \cmidrule(lr){5-6} \cmidrule(lr){7-8}
      & & ASR & ACP & ASR & ACP & ASR & ACP \\
    \midrule
    \multicolumn{9}{l}{\textit{Related Work}} \\
    Diffusion Model & \cite{wewer25srm} & 99.4\% & --$^\star$ & 53.6\% & --$^\star$ & 0.8\% & --$^\star$ & 118M \\
    SRM & \cite{wewer25srm} & 99.8\% & --$^\star$ & 75.4\% & --$^\star$ & 51.6\% & --$^\star$ & 118M \\
    \midrule
    \multicolumn{9}{l}{\textit{NCAs}} \\
    NCA w. 3x3 Conv. & Ours & 99.40 $\pm$ 0.14 & 99.90 $\pm$ 0.00 & 80.90 $\pm$ 0.90 & 98.33 $\pm$ 0.10 & 71.98 $\pm$ 1.61 & 98.20 $\pm$ 0.08 & 1.1M + 50k \\
    NCA w. 5x5 Conv. & Ours & 99.13 $\pm$ 0.50 & 99.87 $\pm$ 0.06 & 79.97 $\pm$ 1.70 & 98.17 $\pm$ 0.15 & 69.00 $\pm$ 6.63 & 95.77 $\pm$ 4.13 & 1.1M + 105k \\
    NCA w. 7x7 Conv. & Ours & 98.73 $\pm$ 0.10 & 99.83 $\pm$ 0.05 & 76.55 $\pm$ 1.21 & 97.73 $\pm$ 0.17 & 72.30 $\pm$ 2.60 & 98.15 $\pm$ 0.19 & 1.1M + 189k \\
    NCA w. 9x9 Conv. & Ours & 99.30 $\pm$ 0.23 & 99.93 $\pm$ 0.05 & 66.73 $\pm$ 16.99 & 95.18 $\pm$ 4.42 & 35.83 $\pm$ 33.56 & 84.25 $\pm$ 20.15 & 1.1M + 300k \\
    NCA w. R/C/B Conv. & Ours & 99.20 $\pm$ 0.08 & 99.90 $\pm$ 0.00 & 80.38 $\pm$ 1.41 & 98.13 $\pm$ 0.17 & 71.25 $\pm$ 2.61 & 98.13 $\pm$ 0.17 & 1.1M + 145k \\
    NCA w. Dilated Perc. & Ours  & 99.25 $\pm$ 0.17 & 99.88 $\pm$ 0.05 & 80.78 $\pm$ 0.95 & 98.13 $\pm$ 0.05 & 70.95 $\pm$ 8.39 & 98.05 $\pm$ 0.57 & 1.1M + 150k \\
    aNCA & Ours & 99.00 $\pm$ 0.37 & 99.88 $\pm$ 0.05 & \textbf{81.90 $\pm$ 5.50} & \textbf{98.40 $\pm$ 0.50} & \textbf{82.10 $\pm$ 4.89} & \textbf{98.85 $\pm$ 0.40} & 1.1M + 165k \\
    \bottomrule
    \multicolumn{9}{l}{\footnotesize $^\star$ indicates metrics not reported in prior work.}\\
  \end{tabular}
\end{table}

\clearpage
\subsection{Maze}
\label{app:maze}

\subsubsection{Model and Training Hyperparameters}
\label{app:hyperparameters_maze}

\begin{table}[h]
  \centering
  \caption{Hyperparameters shared across all NCA variants for the maze task. Perception-specific settings vary per variant and are described in \cref{sec:methods}.}
  \label{tab:hyperparameters_maze}
  \setlength{\tabcolsep}{6pt}
  \begin{tabular}{llr}
    \toprule
    Category & Hyperparameter & Value \\
    \midrule
    \multirow{3}{*}{NCA state}
      & Cell channels & 24 \\
      & Positional embeddings & no \\
      & Fire rate & 0.5 \\
    \midrule
    \multirow{1}{*}{Update network}
      & Architecture & MLP ($1{\times}1$ Conv), 1 hidden layer, 128 channels \\
    \midrule
    \multirow{6}{*}{Optimization}
      & Optimizer & AdamW \\
      & Learning rate & $1\times10^{-3}$ \\
      & Weight decay & $1\times10^{-4}$ \\
      & LR schedule & WSD (warmup--stable--decay) \\
      & Warmup / stable / decay & 10\% / 80\% / 10\% \\
      & Gradient clip norm & 1.0 \\
    \midrule
    \multirow{3}{*}{Training}
      & Total steps & 100\,000 \\
      & Batch size & 64 \\
      & Rollout steps $T$ & $\mathcal{U}[18,\,24]$ \\
    \midrule
    \multirow{2}{*}{Evaluation}
      & Inference rollout steps & 100 \\
      & Path prediction & threshold at $0.5$ (no post-processing) \\
    \midrule
    \multirow{2}{*}{Sample pool}
      & Pool size & 512 \\
      & Pool fraction & 0.5 \\
    \bottomrule
  \end{tabular}
\end{table}

\subsubsection{Training Loss}
\label{app:loss_maze}

Let $\hat{y}_i \in [0,1]$ be the predicted path probability at cell $i$, $y_i \in \{0,1\}$ the ground-truth binary path mask, and $\mathcal{F}$ the set of non-wall cells. Walls are excluded since they are trivially predictable from the condition input. The loss combines three terms. A \textit{BCE} term encourages binary commitment to path vs.\ non-path:
\begin{equation}
  \mathcal{L}_\text{BCE} = -\frac{1}{|\mathcal{F}|} \sum_{i \in \mathcal{F}} \bigl[y_i \log \hat{y}_i + (1-y_i)\log(1-\hat{y}_i)\bigr].
\end{equation}
A \textit{Dice} term directly optimizes path-prediction overlap, handling the imbalance between path and non-path cells:
\begin{equation}
  \mathcal{L}_\text{Dice} = 1 - \frac{2\displaystyle\sum_{i \in \mathcal{F}} \hat{y}_i\, y_i + \epsilon}{\displaystyle\sum_{i \in \mathcal{F}} \hat{y}_i + \sum_{i \in \mathcal{F}} y_i + \epsilon}.
\end{equation}
A \textit{length penalty} penalizes paths that are longer or shorter than the shortest-path target, normalized by the number of free cells:
\begin{equation}
  \mathcal{L}_\text{len} = \left|\frac{\sum_{i \in \mathcal{F}} \hat{y}_i}{|\mathcal{F}|} - \frac{\sum_{i \in \mathcal{F}} y_i}{|\mathcal{F}|}\right|.
\end{equation}
The total loss is $\mathcal{L} = \mathcal{L}_\text{BCE} + \mathcal{L}_\text{Dice} + w_\text{len}\,\mathcal{L}_\text{len}$ with $w_\text{len} = 0.05$.

\clearpage
\subsubsection{Additional Results: Mazes}
\label{app:maze_results}

This section supplements the main maze evaluation with qualitative rollout trajectories and a quantitative same-resolution distribution-shift experiment. We evaluate the trained $30\times30$ aNCA checkpoint, without finetuning, on 1000 unseen Prim-generated mazes and compare its Valid Path and Valid Optimal scores with those on the original Sapient test distribution.

\noindent\textbf{Generalization beyond the training maze distribution.}
To evaluate whether the trained aNCA learns a reusable path-finding procedure, we run the same checkpoint, without finetuning, on 1000 procedurally generated $30\times30$ mazes using randomized Prim-style maze generation~\cite{6773228}. Although these mazes differ from the training distribution in wall density and corridor geometry, the NCA propagates a connected path from start to goal and adapts its rollout to the new layout. It achieves $100\%$ Valid Path and $99.4\%$ Valid Optimal, showing robust transfer to a new maze distribution at the same grid size. Qualitative examples and the full comparison are shown in \cref{fig:maze_generalization_eval}.


\begin{table}[h!]
  \caption{Generalization of the deformable aNCA maze solver to a different maze distribution at the training resolution. The model is trained on $30\times30$ Sapient mazes and evaluated without finetuning on 1000 mazes per test distribution. Valid Path checks connectivity and wall avoidance. Valid Optimal additionally requires shortest-path length.}
  \label{tab:maze_size_generalization}
  \centering
  \begin{tabular}{lcccc}
    \toprule
    Test Maze Size & Generator & Valid Path & Valid Optimal \\
    \midrule
    $30\times30$ & Sapient & 98.56\% & 94.18\% \\
    \midrule
    $20\times20$ & Prim & 100\% & 100\% \\
    $30\times30$ & Prim & 100\% & 99.4\% \\
    \bottomrule
  \end{tabular}
\end{table}
\clearpage

\subsection{Global Color Balance}
\label{app:color_balance}

\subsubsection{Model and Training Hyperparameters}
\label{app:hyperparameters_color_balance}

\begin{table}[h]
  \centering
  \caption{Hyperparameters for the global color balance task.}
  \label{tab:hyperparameters_color_balance}
  \setlength{\tabcolsep}{6pt}
  \begin{tabular}{llr}
    \toprule
    Category & Hyperparameter & Value \\
    \midrule
    \multirow{3}{*}{NCA state}
      & Cell channels & 24 \\
      & Positional embeddings & no \\
      & Fire rate & 0.5 \\
    \midrule
    \multirow{1}{*}{Update network}
      & Architecture & MLP ($1{\times}1$ Conv), 1 hidden layer, 128 channels \\
    \midrule
    \multirow{5}{*}{Optimization}
      & Optimizer & AdamW \\
      & Learning rate & $3\times10^{-4}$ \\
      & LR schedule & WSD (warmup--stable--decay) \\
      & Warmup / stable / decay & 10\% / 80\% / 10\% \\
      & Gradient clip norm & 1.0 \\
    \midrule
    \multirow{3}{*}{Training}
      & Total steps & 25\,000 \\
      & Batch size & 16 \\
      & Rollout steps $T$ & $\mathcal{U}[12,\,24]$ \\
    \midrule
    \multirow{3}{*}{Evaluation}
      & Inference rollout steps & 200 \\
      & Test samples per run & 1000 \\
      & Independent training runs & 3 \\
    \midrule
    \multirow{2}{*}{Sample pool}
      & Pool size & 512 \\
      & Pool fraction & 0.5 \\
    \midrule
    \multirow{2}{*}{Loss weights}
      & $w_\text{count}$ & 1.0 \\
      & $w_\text{sat}$ & 0.01 \\
    \bottomrule
  \end{tabular}
\end{table}

\subsubsection{Training Loss}
\label{app:loss_color_balance}

Let $x_i \in [-1, 1]$ be the predicted pixel value of cell $i$ (where $-1$ = black, $+1$ = white), $N$ the total number of pixels, and $\mu^* = 2r - 1$ the target mean corresponding to a prescribed white fraction $r$. The loss combines two terms. A \textit{count term} pushes the global pixel mean toward the target ratio:
\begin{equation}
  \mathcal{L}_\text{count} = \left(\frac{1}{N}\sum_{i=1}^{N} x_i - \mu^*\right)^2.
\end{equation}
A \textit{saturation term} pushes each pixel to commit to either $-1$ or $+1$, zeroing only at fully binary values:
\begin{equation}
  \mathcal{L}_\text{sat} = \frac{1}{N}\sum_{i=1}^{N} \left(1 - x_i^2\right).
\end{equation}
The total loss is $\mathcal{L} = \mathcal{L}_\text{count} + \mathcal{L}_\text{sat}$.
\newpage
\subsubsection{Additional Results: Global Color Balance}
\label{app:color_balance_results}

\begin{figure}[h!]
  \centering
  \includegraphics[width=0.7\linewidth]{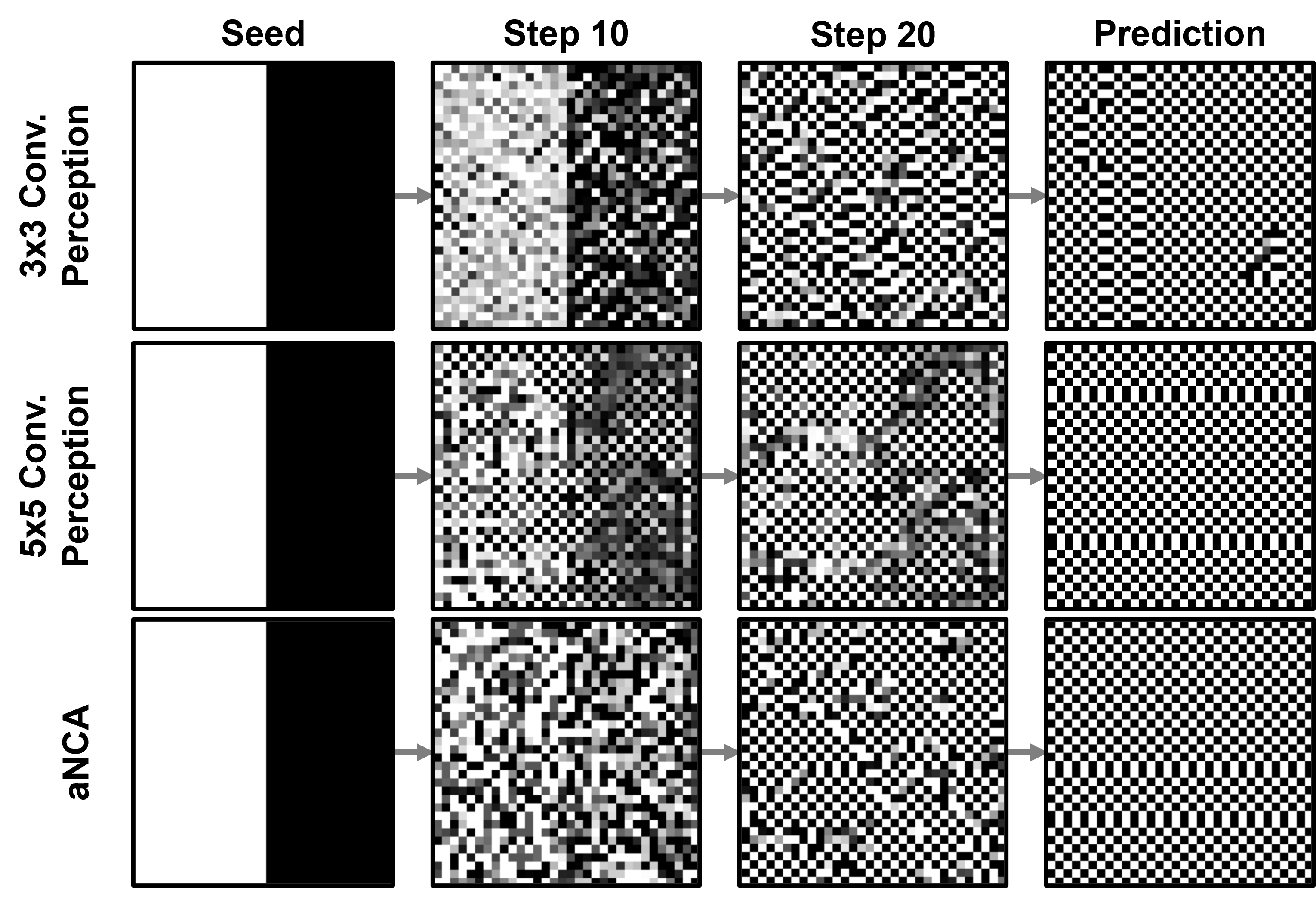}
  \caption{Global color balance: the model must satisfy a prescribed black/white ratio without per-pixel supervision.}
  \label{fig:color_balance}
\end{figure}

\end{document}